%% file: main.tex
\documentclass[twocolumn]{fairmeta}
\usepackage{soul}

\usepackage{xspace}  % for macros to add space where needed 
\makeatletter
\DeclareRobustCommand\onedot{\futurelet\@let@token\@onedot}
\def\@onedot{\ifx\@let@token.\else.\null\fi\xspace}
\makeatother 
\def\eg{\emph{e.g}\onedot}

\newcommand{\benchmark}{DIMCIM\xspace}
\newcommand{\cocobenchmark}{COCO-DIMCIM\xspace}
\newcommand{\doesit}{Does-It Metric\xspace}
\newcommand{\canit}{Can-It Metric\xspace}
\newcommand{\MetaLDM}{Flow-Int\xspace}
\newcommand{\sdtwo}{LDM2.1\xspace}
\newcommand{\sdthree}{LDM3.5L\xspace}
\newcommand{\flux}{FLUX.1-dev\xspace}

\usepackage{listings}

\title{DIMCIM: A Quantitative Evaluation Framework for Default-mode Diversity and Generalization in Text-to-Image Generative Models}

\author[1,2]{Revant Teotia}
\author[1]{Candace Ross}
\author[1]{Karen Ullrich}
\author[2]{Sumit Chopra}
\author[1,3,4,5]{Adriana Romero-Soriano}
\author[1,*]{Melissa Hall}
\author[1,*]{Matthew Muckley}

\affiliation[1]{FAIR at Meta - New York and Montreal labs}
\affiliation[2]{Courant Institute for Mathematical Sciences, NYU}
\affiliation[3]{Mila - Quebec AI Institute}
\affiliation[4]{McGill University}
\affiliation[5]{Canada CIFAR AI chair}

\contribution[*]{Joint last author}

\abstract{\input{sec/0_abstract}}
 
\date{\today}

\correspondence{Revant Teotia at \email{rt2741@nyu.edu}}
\metadata[Data and code]{\url{https://github.com/facebookresearch/DIMCIM}}
\begin{document}

\maketitle

\input{sec/1_intro}

\input{sec/2_relatedwork}
\input{sec/3_benchmark}
\input{sec/4_analysis}
\input{sec/6_discussion}

\clearpage
\newpage
\bibliographystyle{assets/plainnat}
\bibliography{main}

\clearpage
\newpage
\onecolumn
\beginappendix
\input{sec/7_appendix}

\end{document}

%% file: sec/0_abstract.tex
Recent advances in text-to-image (T2I) models have achieved impressive quality and consistency.
However, this has come at the cost of representation diversity.
While automatic evaluation methods exist for benchmarking model diversity, they either require reference image datasets or lack specificity about the kind of diversity measured, limiting their adaptability and interpretability.
To address this gap, we introduce the Does-it/Can-it framework, \benchmark, a reference-free measurement of default-mode diversity (``Does" the model generate images with expected attributes?) and generalization capacity (``Can" the model generate diverse attributes for a particular concept?).
We construct the \cocobenchmark benchmark, which is seeded with COCO concepts and captions and augmented by a large language model. 
With \cocobenchmark, we find that widely-used models improve in generalization at the cost of default-mode diversity when scaling from 1.5B to 8.1B parameters.
\benchmark also identifies fine-grained failure cases, such as attributes that are generated with generic prompts but are rarely generated when explicitly requested. 
Finally, we use \benchmark to evaluate the training data of a T2I model and observe a correlation of 0.85 between diversity in training images and default-mode diversity.
Our work provides a flexible and interpretable framework for assessing T2I model diversity and generalization, enabling a more comprehensive understanding of model performance.

%% file: sec/1_intro.tex
\section{Introduction}
\label{sec:intro}

In recent years, text-to-image (T2I) generative models have witnessed impressive advances, yielding unprecedented photorealistic quality. These advances have been driven by models optimized for human preference. 
Performance improvements for these models are usually reported in terms of image realism and prompt-image consistency, with well-established metrics~\citep{heusel2018FID, kynkäänniemi2019improvedprecisionrecallmetric, naeem2020reliablefidelitydiversitymetrics_coveragedensity, hessel-etal-2021-clipscore, hu2023tifaaccurateinterpretabletexttoimage_tifa, cho2024davidsonianscenegraphimproving_DSG, lin2024evaluating_vqascore}.
However, optimizing for human preference alone has led to state-of-the-art models having limited representation diversity~\citep{hall2024digin,sehwag2022generating,zameshina2023diverse,corso2023particle,askari2023feedback,hemmat2024improvinggeodiversitygeneratedimages,sadat2023cads}.

There have been attempts to quantify such diversity challenges, for example, in~\citep{dincà2024openbiasopensetbiasdetection}, which focuses on the bias of images generated in default settings.
However, these methods %may not correspondingly
do not take into account model generalization capacity, i.e., whether models can generate fully diverse sets of images through prompt adjustments.
Furthermore, \emph{reference-free} diversity metrics (like Vendi Score~\citep{friedman2023vendiscorediversityevaluation_vendiscore}) do not measure how well generated images capture definitions of diversity relevant to the real world.
On the other hand, \emph{reference-based} (FID~\citep{heusel2018FID}, Recall \citep{kynkäänniemi2019improvedprecisionrecallmetric}, Coverage \citep{naeem2020reliablefidelitydiversitymetrics_coveragedensity}) require curating new image datasets for every new set of categories of interest, making them less flexible, expensive and difficult to adapt.
Finally, existing metrics usually present a single summary statistic of diversity with limited fine-grained information and are not very interpretable or actionable.

\begin{figure}[t]
  \centering
  \includegraphics[width=\linewidth]{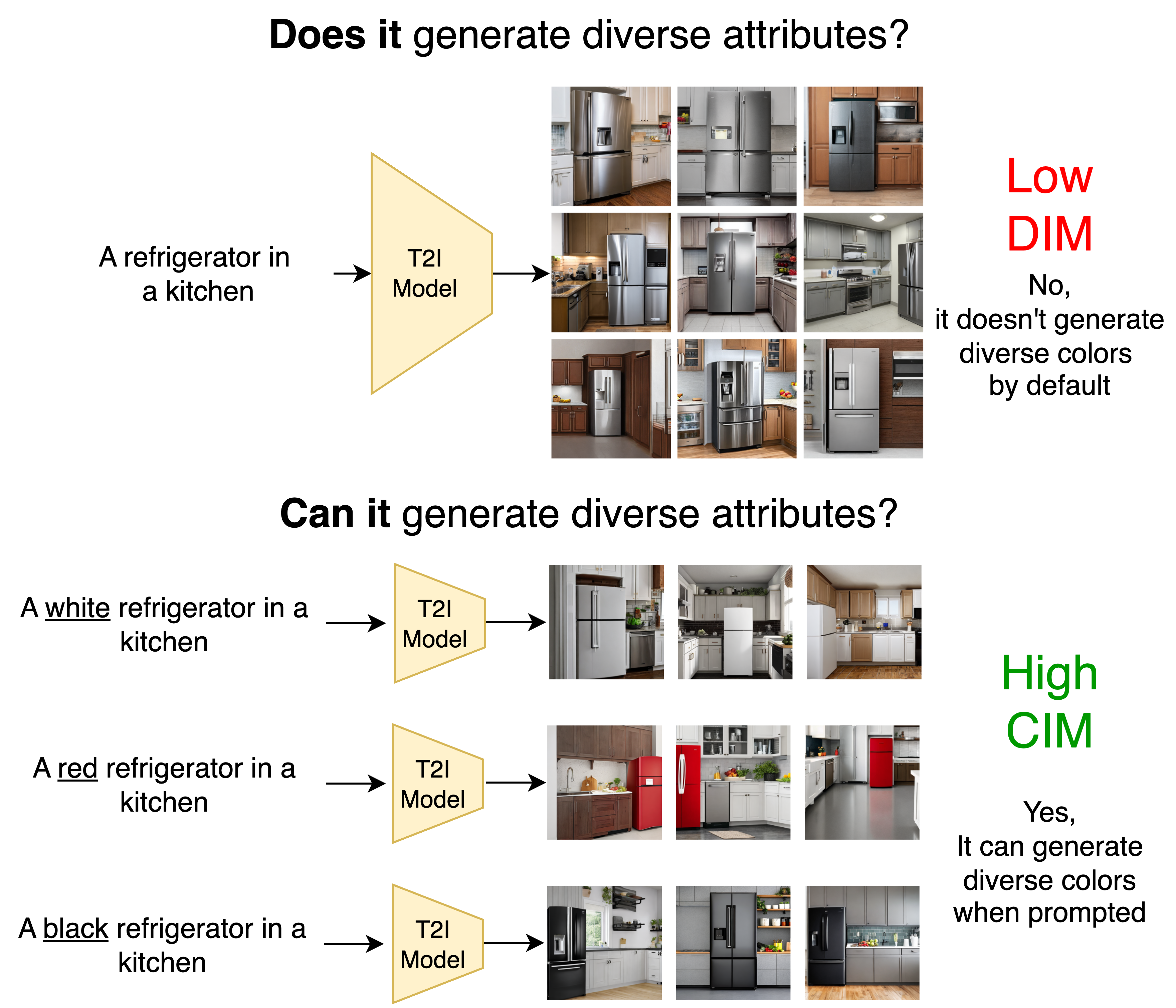}
  \caption{
  \textbf{\benchmark can be used to measure default-mode diversity (\doesit) and generalization capabilities (\canit) of text-to-image models.}
  For example, \sdtwo ~\citep{Rombach_2022_CVPR_SD2_1} has a low ~\doesit score for the ``color'' attribute type and does not generate diverse colored refrigerators for generic input prompts, revealing default-mode diversity challenges. 
  However, the model has a high \canit when explicitly prompted with different colors, revealing strong generalization capabilities. 
  }
  \label{fig:summary_figure}
\end{figure}

To address these limitations, we introduce a reference-free benchmarking framework that provides fine-grained information about image generation diversity and model generalization capacity.
Our method measures two aspects of image diversity within T2I generative models (see ~\Cref{fig:summary_figure}): 1) Without explicit prompting, \emph{does} the model generate images with a variation of expected attributes? and 2) With explicit prompting, \emph{can} the model generate diverse attributes for a particular concept? 
The first measurement corresponds to default-mode diversity, and we call it the \textbf{Does-it Metric} (DIM).
The second measurement corresponds to generalization capacity, and we call it the \textbf{Can-it Metric} (CIM).
We call our framework \textbf{\benchmark}.

\benchmark is constructed hierarchically based on concepts (e.g., ``car'', ``refrigerator'', or ``dog'') and paired attributes (e.g., ``red'', ``closed'', or ``terrier'') from a pre-existing reference prompt dataset.
We use an LLM to generate two sets of T2I generation prompts from the reference prompts: 1) under-specified \textit{coarse prompts} with attribute information removed and 2) \textit{dense prompts} with diverse attributes explicitly added to concepts in coarse prompts. We then use a modified procedure of the VQAScore~\citep{lin2024evaluating_vqascore} to identify concept-attribute combinations in images and compute the \doesit and \canit using coarse and dense prompts, respectively. 
Lastly, we develop a simple summarization statistic and normalization procedure to allow comparisons across concepts, attributes, and models.

To summarize, we enumerate our contributions as:
\begin{itemize}
    \item We propose \benchmark, a new quantitative assessment for both \textit{default-mode diversity} (the \doesit) and \textit{generalization capacity} (the \canit) of text-to-image generative models, outlined in \Cref{sec:benchmark_framework}.
    
    \item Demonstrating the flexibility and adaptability of \benchmark, we develop the \cocobenchmark benchmark based on objects in the COCO~\citep{lin2015microsoftcococommonobjects_coco} dataset and utilize Llama3.1~\cite{grattafiori2024llama3herdmodels} to build prompts that pertain to real-world diversity, explained in \Cref{sec:benchmark_dataset}.
    
    \item Using \cocobenchmark, we perform an analysis of existing state-of-the-art text-to-image generative models and identify a notable trade-off in default-mode diversity and generalization capabilities, described in \Cref{sec:experiments_overall}. 
    
    \item We utilize fine-grained information from \benchmark to identify failures of open-source models, including expected failures (e.g., negation) and unexpected failures (e.g., ``closed refrigerator''), described in \Cref{subsec:finding_failure_modes}.
    
    \item Using a proprietary model with known training data, we identify strong correlations between the visual attributes of the training data and the model's default-mode diversity abilities while also demonstrating the utility of \benchmark in identifying anomalous cases where increased representation in training data images \textit{does not} coincide with improved diversity in generated images. This is described in~\Cref{subsec:training_data_investigation}.
\end{itemize}

%% file: sec/2_relatedwork.tex
\section{Related Work}
\label{sec:relatedwork}
\if0
Outline:
\begin{itemize}[noitemsep,nolistsep]
    \item[--] generation diversity metrics
    \item[--] T2I consistency metrics (CLIPScore, TIFA, DSG, VQAScore)
    \item[--] validity of using LLMs to generate text prompts (T2I-Compbench)
    \item[--] datasets to measure diversity/ability to compose concepts for generation (Winoground-T2I, ConceptMix, that paper I reviewed from ACL last year on people diversity -- kind of meh tbh can't remember the name)
    \item[--] (maybe include) challenges of measuring diversity in static datasets
\end{itemize}
\fi

Existing approaches to measure the diversity of generated images can be categorized as \emph{reference-free} metrics and \emph{reference-based} metrics. 
The reference-free metrics, such as the Vendi Score \citep{friedman2023vendiscorediversityevaluation_vendiscore}, computes the pairwise similarity within a batch of samples, without depending on any dataset. 
One weakness of this metric is that pairwise similarity does not capture real-world diversity. 
On the other hand, the \emph{reference-based} metrics, \eg FID \citep{heusel2018FID}, recall \citep{kynkäänniemi2019improvedprecisionrecallmetric} and coverage \citep{naeem2020reliablefidelitydiversitymetrics_coveragedensity}, depend on some datasets typically constituting real-world images. 
Their primary weakness is that any axis of diversity, \eg color or shape attributes, requires a large set of comparison images with these attributes.
Concurrent to our work, GRADE \citep{rassin2025grade} measures the distribution of generated attributes for a given concept as a proxy for diversity.
This explores default mode bias (\eg given the prompt ``image of a cookie,'' how diverse are the generations?) but not generalization capacity (\eg when prompted ``image of a star-shaped cookie,'' can the model do it?).

In addition to metrics, there are existing prompt datasets for evaluating how well models can compose attribute-object combinations. These datasets are either template-based 
(Winoground-T2I \citep{zhu2023contrastive}, T2I-Compbench \citep{huang2023t2i}, ABC-6K \citep{feng2022training}) or fully free-form using an LLM (ConceptMix \citep{wu2024conceptmix}).
While these datasets also often leverage an LLM for prompt generation, one key difference is the degree of attribute coverage. 
These works may test a few attribute-object combinations (\eg \textit{red dog} and \textit{blue dog}), whereas we extensively generate a large number of prompts and attributes for a given concept. 

Lastly, there are many consistency metrics \citep{hessel-etal-2021-clipscore,cho2023visual,cho2023davidsonian,tan2024evalalign} that evaluate the faithfulness of generated images to text prompts. Our framework leverages these metrics to evaluate a large number of attribute-object pairs. Some metrics \citep{hu2023tifa,cho2023davidsonian,cho2023visual} generate specific questions given the prompt (\eg \textit{Is there a dog to the left of the yellow couch?}), then evaluate these questions using a VQA model. 
One drawback is that these metrics require multiple systems (at a minimum, an LLM for question generation and a VQA model for question evaluation) and are susceptible to linguistic biases \citep{ross2024makes} and poorly generated questions for more complex prompts \citep{lin2024evaluating_vqascore}. 
VQAScore \citep{lin2024evaluating_vqascore} does not suffer from the need for an external LM, yet still uses the entire prompt for evaluation. We use a modification of the VQAScore metric, where we only consider the specific \textit{attribute-object} at hand.

%% file: sec/3_benchmark.tex
\section{\benchmark Framework}
\label{sec:benchmark_framework}

We present the \benchmark framework, a quantitative benchmarking method that allows for actionable insights about image generation diversity.
We focus on a concept's real-world visual diversity, such as variations in color, type, material, or size. 
Given any T2I model, our benchmarking method focuses on two diversity-related objectives:
\begin{enumerate}
    \item Default-mode Diversity: Here we ask the question ``\textit{Does} the model generate images of concepts with visual diversity in its default setting with general prompts that do not contain explicit diversity-related instructions?'' 
    We measure this property with our \doesit.
    \item Model Generalization: Under this objective, we ask ``\textit{Can} the model generate images of concepts with diverse attributes when explicitly prompted?''
    We measure this property with our \canit.
\end{enumerate}

\subsection{Preliminary Definitions}

We ground our discussion by first providing some definitions. 
\benchmark is built to assess how well a text-to-image generative model depicts the diversity (either by default or with explicit prompting) of a set of \textbf{concepts} $C$.
For example, a set $C$ of possible concepts $c$ could be $C = \{dog, hat, car\}$.
Each concept $c$ has a corresponding set of \textbf{attribute types} $T_c$.
Each attribute type is an axis of diversity relevant to that concept. 
For example, the concept $dog$ could contain attribute types $T_{dog} = \{color, breed, position\}$ while the concept $hat$ could contain attribute types $T_{hat} = \{color, style, material\}$. 
Each attribute type has a corresponding set of \textbf{attributes} $A_{c,t}$.
For example, the attribute type of $dog:color$ could contain attributes $A_{dog,color} = \{black, white, golden\}$ while the attribute type $hat:color$ could contain attributes $A_{hat,color} = \{black, pink, rainbow\}$

Given a set of concepts, attribute types, and attributes, we construct prompts for benchmarking. 
We define a \textbf{coarse prompt} as a prompt $p_{c,-}$ that contains a single target concept $c$ without mention of visual attributes.
An example coarse prompt for the concept $dog$ could be $p_{dog,-} = $ ``The dog plays in the grass.''
Using coarse prompts, we can identify whether a model generates images of a given concept with visual diversity by default without explicit attribute information. 
A \textbf{dense prompt} is a prompt $p_{c,a}$ that contains concept $c$ and attribute $a \in A_{c,t}$. 
We call this a ``dense'' prompt because it is an augmentation of the original coarse prompt $A_{c,-}$ with additional attribute information.
This allows us to understand whether the model is capable of generating % various types of 
visual diversity when explicitly requested.
An example dense prompt for the previous coarse prompt would be $p_{dog,brown}$ = ``The brown dog plays in the grass.''

\begin{figure*}[h]
  \centering
  \includegraphics[width=\textwidth]{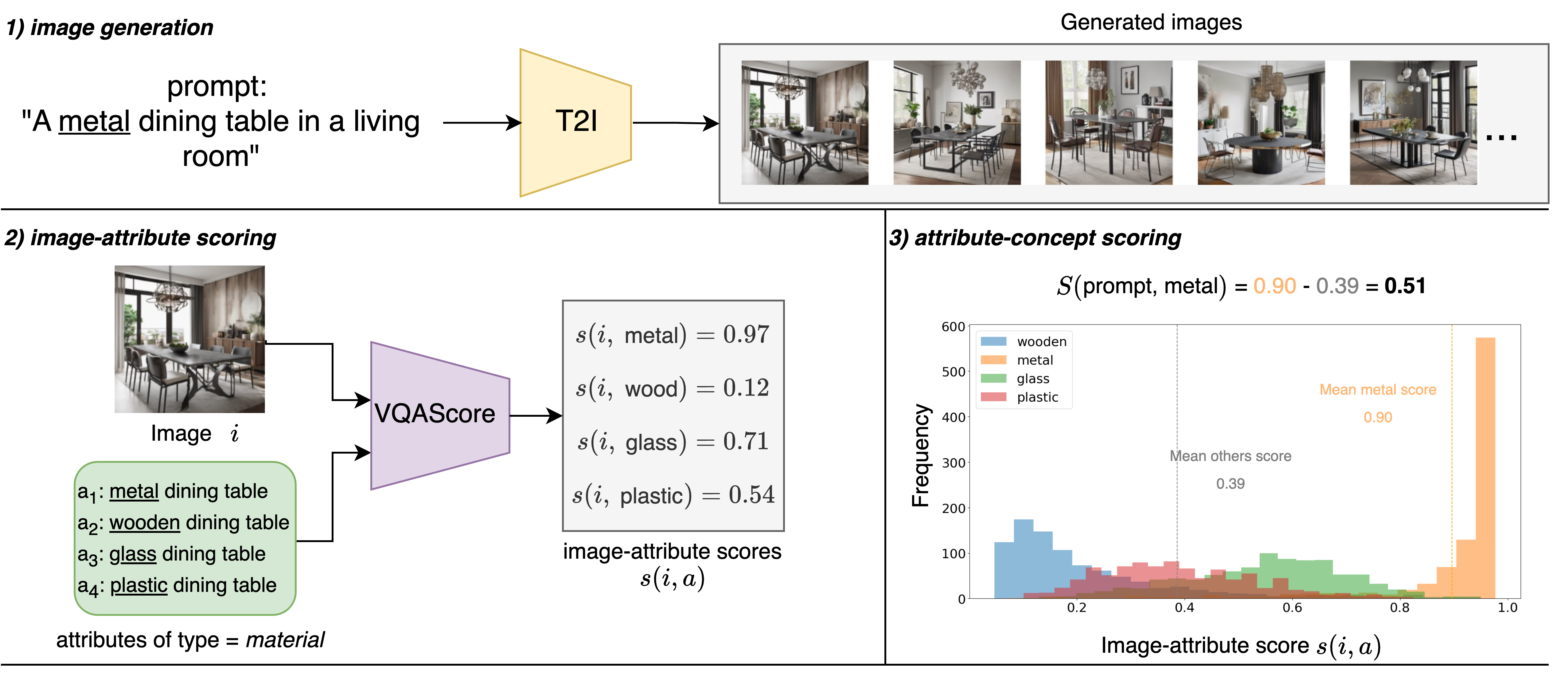}
  \caption{
  \textbf{An example of attribute-concept scoring in \benchmark.}
   To calculate the attribute-concept score  corresponding to the concept \textit{table}, attribute type \textit{material}, and attribute ``metal'' with prompt ``A  metal dining table in a living room," there are three steps:
    \textbf{[Top]} First, generate images.  
    \textbf{[Left]} Then, for each image $i$, calculate the image-attribute score $s(i,a)$ for each candidate attribute $a$ using the VQAScore~\citep{lin2024evaluating_vqascore} 
    \textbf{[Right]} Finally, aggregate all the image-attribute scores, where the attribute-concept score for ``metal'' $S(prompt, metal)$ is the difference between the mean ``metal'' scores of all images and the mean of all other attribute scores.
  }
  \label{fig:score_calculation}
\end{figure*}

\subsection{Attribute-Concept Scoring}
\label{subsec:attribute_concept_scoring}
To quantify the notion of diversity of any T2I model, we first need to quantify how much any attribute $a$ is preferred over the other attributes in $A_{c,t}$ in the model generated images for an input prompt $p_c$.
For that, we introduce the notion of ``attribute-concept score'' $S(p_c,a)$ for a
given prompt $p_c$ of concept $c$ and an attribute $a \in A_{c, t}$ of attribute type $t$. 

To calculate $S(p_c,a)$, we first generate $n$ images for the input prompt $p_c$. Then for each generated image $i$ we calculate the presence of all attributes $a_j \in A_{c,t}$ in the image as image-attribute score $s(i,a_j)$. We use Vision Language Models (VLMs) to calculate $s(i,a_j)$. More specifically, we use VQAScore~\citep{lin2024evaluating_vqascore} which uses a visual-question-answering model to produce an alignment score corresponding to the probability of a ``Yes'' answer to the question, ``\texttt{Does this figure show \{text\}?}'', where \texttt{\{text\}} corresponds to the given concept $c$ and attribute $a$ combination. However, any of the SOTA VLMs that give image-text alignment scores (like \cite{hessel-etal-2021-clipscore}) can be used to calculate $s(i,a_j)$, so we chose VQAScore because it has been shown to better align with human judgment in ~\citep{lin2024evaluating_vqascore}. During experiments, we found that using full prompt as VQAScore input \texttt{\{text\}} causes image background context to influence the image-attribute score $s(i,a_j)$. For example, if we use ``A metal dining table in a living room", the ``living room'' in image also contributes to $s(i,a_j)$. We therefore use truncated text, like ``a metal dining table," containing only concept and attribute as VQAScore input \texttt{\{text\}} to minimize the influence of image background that are not relevant to concept attributes. 

Once we have calculated image-attribute scores ($s(i,a_j)$) for all the $n$ images, we define the ``attribute-concept score'' $S(p_c,a)$ as the difference between the mean image-attribute score of attribute $a$ and the mean image-attribute score of all the other attributes in $A_{c,t}$. i.e. 
\begin{equation}
  S(p_c,a) = \frac{\sum{s(i,a)}}{n} - \frac{\sum{s(i,a_j)}}{n(|A_{c,t}| - 1)}
  \label{eq:attribute_preference_score}
\end{equation}
where $a_j \in A_{c,t}$ and $a_j \neq a$. See \Cref{fig:score_calculation} for an example.

\subsection{Summary Statistics}
\label{subsec:summary_stat}
We then aggregate these scores to build our \doesit and \canit to provide insights into the default bias and generalization capabilities, respectively, of T2I models. 

\vspace{1mm}

\noindent \textbf{\doesit}
We calculate a model's default mode diversity using coarse prompts. 
For a coarse prompt $p_{c,-}$ where no attribute is specified for the target concept $c$ in the prompt, the attribute-concept score $S(p_{c, -},a)$ corresponds to the representation of the model for that attribute $a$ compared to other attributes of the same type $A_{t,c}$. 
A large positive (or negative) value of $S(p_{c, -},a)$ indicates that attribute $a$ is generated more (or less) than other attributes of the same type in the images generated by the coarse prompt. 
We use $S(p_{c, -},a)$ as the \doesit score at attribute level. 
A higher attribute level \doesit means that, in default-mode, the attribute is generated more than other attributes of the same type with lower \doesit .
We report 1 minus the mean of the absolute value of $S(p_{c, -},a)$ across all concepts and attributes as the summary \doesit, where the absolute value of $S(p_{c, -},a)$ indicates the extent of imbalance in the model's depiction of attribute $a$.
Thus, a high summary \doesit means the model generates images with a balanced representation across attributes and is thus more diverse, while a low score means the generations are highly imbalanced.

\vspace{1mm}

\noindent \textbf{\canit} 
We calculate a model's generalization capacity using dense prompts. 
For a dense prompt $p_{c,a}$ where the attribute $a \in A_{t,c}$ of concept $c$ is explicitly specified in the prompt, we use the attribute-concept score $S(p_{c, a},a)$ as an attribute level \canit to indicate a model's capacity to generate the attribute $a$. 
$S(p_{c, a},a)$ 
measures how much the selected attribute $a$ is represented in the generated images compared to other attributes of the type $A_{c,t}$.
For images generated with prompt $p_{c, a}$, a high $S(p_{c, a},a)$ means that attribute $a$ is generated more than other attributes of the same type.  
While a low $S(p_{c, a},a)$ means that attribute $a$ is generated less than other attributes of the same type, even though it is specifically requested. 
We define the summary \canit as the mean of $S(p_{c, a},a)$ for all $p_{c, a}$.

\section{\cocobenchmark Dataset}
\label{sec:benchmark_dataset}

In this Section we introduce the \cocobenchmark Dataset, a benchmarking dataset of concepts, attributes, coarse prompts and dense prompts derived from COCO~\cite{lin2015microsoftcococommonobjects_coco} by leveraging the \benchmark framework described in ~\cref{sec:benchmark_framework}. 

\vspace{1mm}

\noindent \textbf{Concepts and attribute collection} % \matt{Break up this paragraph} 
We use the COCO~\cite{lin2015microsoftcococommonobjects_coco} dataset which contains images of every-day, common objects with human written captions.
We select 30 object classes to build the concept set $C$. 
For each concept $c$, we randomly select 31 COCO captions as seed prompts $p_c$. 
Each seed prompt contains the selected concept as its main subject, although there may be other auxiliary nouns in the seed prompts too. We used spaCy~\cite{Honnibal_spaCy_Industrial-strength_Natural_2020} to find captions that have the selected concept as their first noun and filtered the ones which had a different main subject.
Note that we filtered out human-related captions and concepts, i.e. captions with words ``child,'' ``person,'' ``woman,'' ``man,'' etc., as our work focuses on visual attribute diversity in generated images of everyday object classes and does not address the question of normative diversity in the depiction of people.

We first use an LLM (Llama3.1~\citep{grattafiori2024llama3herdmodels}) to collect a set of candidate attribute types and attributes for each concept.
We do so by passing all COCO seed prompts $p_c$ for a concept $c$ through the LLM and extracting text that corresponds to visual descriptions of concept $c$.
Additionally, we use the LLM to generate additional visual attributes that could be plausible for concept $c$ in the context of prompt $p_c$.
Using multiple seed prompts corresponding to naturally occurring everyday images ensures that generated attributes are plausible in the real world (not something like ``a rainbow colored dog on Mars'') while still diverse.

We then ask the LLM to group attributes for each concept $c$ by attribute type $T_c$.
For example ``black,'' ``brown,'' and ``wood'' are grouped as \textit{color} while ``wood,'' ``plastic,'' and ``metal'' are grouped as \textit{material} for the concept \textit{table}. 
For a concept $c$, we manually filter out attributes that are visually ambiguous/indistinguishable in images (like age, motion, size, model name, accessories, etc.) or are mutually non-exclusive. Meta-prompts used for instructing the LLM can be found in the Supplementary Materials.

\vspace{1mm}

\noindent \textbf{Coarse and dense prompts collection} 
We use the COCO seed prompts to generate coarse prompts by asking the LLM to remove existing visual attribute (if any) about the concept and rewrite the prompt while preserving the context/environment. 
For example,  for seed prompt ``A \underline{wooden} \textit{table} in a living room,'' the generated coarse prompt is ``A \textit{table} in a living room.'' 
Preserving context from the seed prompts ensures diversity of object environment while maintaining plausibility. A figure for prompt construction pipeline can be found in the Supplementary Material.

We generate dense prompts by injecting visual attributes of the concepts into coarse prompts. For each coarse prompt, we use the LLM to inject one attribute at a time to the main subject in the coarse prompt to make a naturally plausible sentence. 
For example the coarse prompt ``A \textit{table} in a living room" is augmented by adding different colors and materials. 
We ask the LLM to skip a particular attribute if injecting it into a coarse prompt results in an unnatural dense prompt, \textit{e.g.} ``A bird \textit{perched} in the air."

This way, we develop a benchmark that consists of 30 concepts, 494 attributes, 930 coarse prompts, and 14,641 dense prompts. Each concept has an average of 4.83 attribute types and 16.46 attributes. 
We generate 31 coarse prompts and an average of 488 dense prompts (15.74 on average per coarse prompt) for each concept.

%% file: sec/4_analysis.tex
\section{Experiments}
\label{sec:analysis}

We use \benchmark to study the default-mode diversity and generalization capabilities of state-of-the-art T2I models. 

We evaluate two generations of a latent diffusion model: ``\sdtwo''~\citep{Rombach_2022_CVPR_SD2_1}, trained on a public dataset of approximately 5 billion images then further trained on images of higher resolution and fine-tuned on aesthetic images, and ``\sdthree'', a large multi-modal diffusion transformer with two text encoders~\citep{esser2024scalingrectifiedflowtransformers_SD3_5}.
We also evaluate with ``\flux''~\citep{flux2024}, a rectified flow transformer with 12 billion parameters.
Finally, we evaluate a proprietary text-to-image model that leverages flow matching~\citep{lipman2023flowmatchinggenerativemodeling_flow_matching}, control conditions~\citep{berrada2025boosting,podell2023sdxlimprovinglatentdiffusion_SDXL} and latent perceptual loss~\citep{berrada2025boosting}.
We denote this model as ``\MetaLDM'' and describe its training details in the Supplementary Material.

For each model, we generate $n=30$ images per dense prompt $p_{c,a}$ and an equivalent number of images for the corresponding coarse prompts $p_{c,-}$ so that there is the same quantity of coarse and dense prompts. 
Following the \benchmark protocol, we use the coarse prompts for calculating the \doesit and the dense prompts for calculating the \canit.
We then compute the \doesit and \canit using VQAScore~\citep{lin2024evaluating_vqascore} as described in ~\Cref{subsec:attribute_concept_scoring} and~\cref{subsec:summary_stat}, using the \textit{InstructBLIP-Flan-T5-XXL} ~\citep{dai2023instructblipgeneralpurposevisionlanguagemodels_instblip, chung2022scalinginstructionfinetunedlanguagemodels_flant5} model\footnote{
We also perform a cross-validation study with other VQAScore models in ~\citet{lin2024evaluating_vqascore} and found that our findings are consistent across different models.}.
For the crux of our analysis, all images are generated with classifier-free guidance scale of 7.5. We also include an analysis across guidance scales.

\subsection{Overall \benchmark results}
\label{sec:experiments_overall}
\begin{table}[h]
  \centering
  \begin{tabular}{lccc}
    \toprule
    Model (size) & DIM & CIM \\
    \midrule
    \sdtwo (1.5B)~\citep{Rombach_2022_CVPR_SD2_1} & 0.815 & 0.299 \\
    \MetaLDM (1.9B) & 0.802 & 0.315 \\
    \sdthree (8.1B)~\citep{esser2024scalingrectifiedflowtransformers_SD3_5} & 0.799 & 0.374 \\
    FLUX.1-dev (12B)~\citep{flux2024} & 0.785 & 0.326 \\
    \bottomrule
  \end{tabular}
  \caption{\textbf{\benchmark identifies a trade-off between default-mode diversity and generalization capability.} As model size increases, the \doesit decreases while \canit increases. 
  } 
  \label{tab:model_comparison}
\end{table}

\begin{table}[h]
  \centering
  \begin{tabular}{lccc}
    \toprule
    Model (CFG scale) & DIM & CIM \\
    \midrule
    \MetaLDM (2.0 CFG) & 0.839 & 0.261 \\
    \MetaLDM (5.0 CFG) & 0.809 & 0.309 \\
    \MetaLDM (7.5 CFG) & 0.802 & 0.315 \\
    \bottomrule
  \end{tabular}
  \caption{
  \textbf{Increasing classifier-free guidance improves generalization capacity makes but decreases their default-mode diversity.} This is likely explained by stronger guidance contributing to greater prompt-image consistency.
  }
  \label{tab:metaldm_cfg_comparison}
\end{table}

\begin{figure*}[h]
  \centering
  \includegraphics[width=\textwidth]{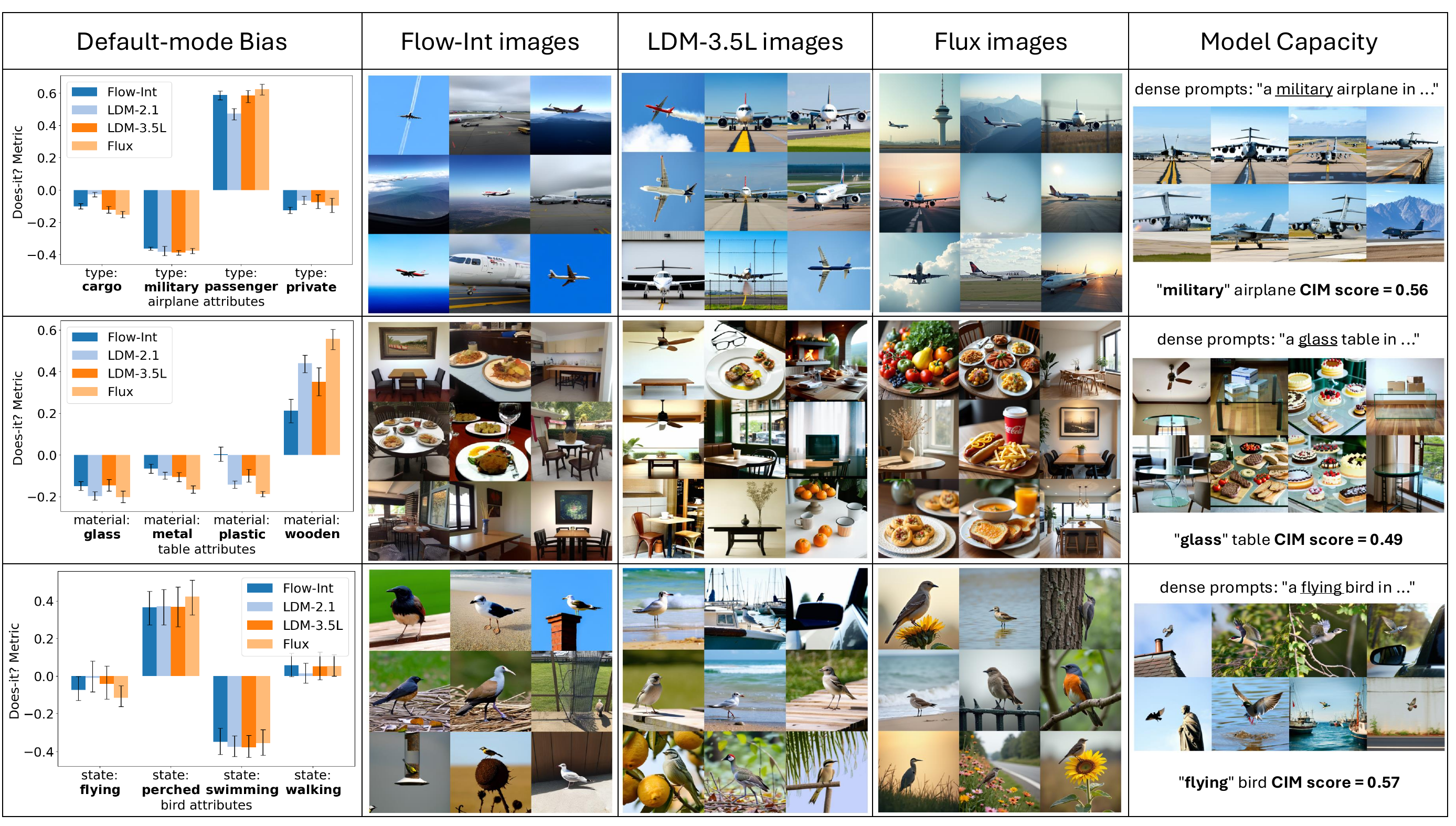}
  \caption{
  \textbf{The \benchmark identifies default-mode diversity limitations with the \doesit. These occur even though the model is capable of generating diverse attributes, identified with the \canit.}  
  \textbf{[Left]} Some concepts imbalanced \doesit scores, revealing default-mode diversity issues. 
  \textbf{[Middle]} These diversity challenges are reflected in random visual examples of coarse prompts, with default towards, \textit{e.g.}, ``passenger'' \textit{airplane}, ``wooden'' \textit{table}, and ``perched`` \textit{bird} across all models. 
  \textbf{[Right]} Using the \canit, we find that some models have the generalization capacity to generate attributes that are under-represented by default when explicitly prompted with dense prompts, including ``military'' \textit{airplane}, ``glass'' \textit{table}, and ``flying'' \textit{bird} (example images from \sdthree).
  }
  \label{fig:default_mode_bias_analysis}
\end{figure*}

We now discuss results using \cocobenchmark, which are summarized in ~\Cref{tab:model_comparison}.
We observe that for the \doesit smaller models (\sdtwo and \MetaLDM) have higher scores than the larger models, \sdthree and \flux.
This means that, when generating images from coarse prompts containing a given concept, smaller models depict a more balanced distribution of attributes for that concept and thus have more diversity in their default mode operation.
On the other hand, the larger models (\sdthree and \flux) have high \canit scores, meaning that they have  a higher proportion of images that contain the desired attribute-concept combination when explicitly prompted to do so, and thus stronger generalization capabilities, while smaller models (\sdtwo and \MetaLDM) have lower \canit scores.
This highlights a notable trade-off in default-mode diversity and generalization capabilities in existing state-of-the-art text-to-image generative models.
We also studied the effect of classifier free guidance (CFG)~\cite{ho2022classifierfreediffusionguidance_CFG}. Increasing CFG increases model generalization capacity while worsening its default mode bias (\Cref{tab:metaldm_cfg_comparison}).

% -----------------

\subsection{Finding model failure modes}
\label{subsec:finding_failure_modes}
We continue our analysis of T2I models to show how model failure modes discovered through \benchmark are easily interpretable: plotting and analyzing the \doesit and \canit for concept-attribute pairs allows useful identification of failure modes and provides actionable insights.

\begin{figure*}[t]
  \centering
  \includegraphics[width=\textwidth]{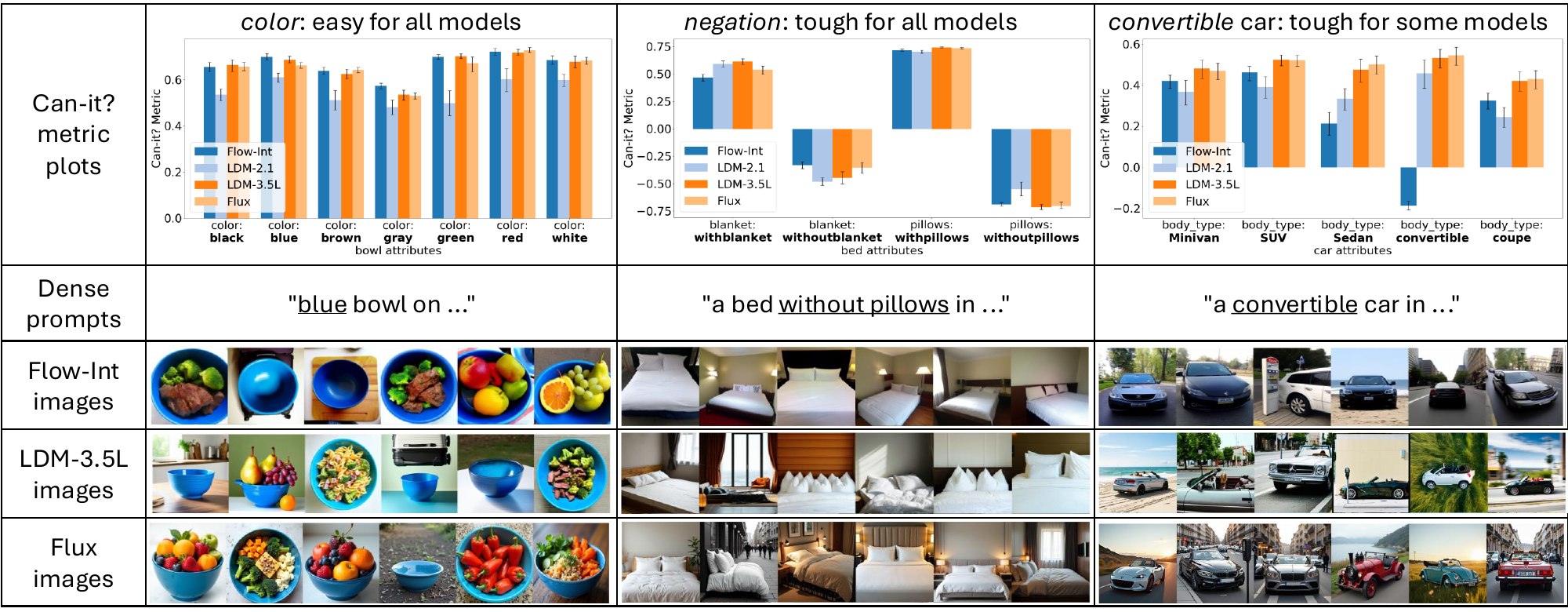}
  \caption{
  \textbf{The \canit identifies model generalization failures.} 
  \textbf{[Left]} Most models are able to generalize across colors.
  \textbf{[Middle]} Models struggle with negation. 
  \textbf{[Right]} \MetaLDM struggles with generating  ``convertible'' \textit{cars} while the other models don't. 
  }
  \label{fig:capacity_analysis}
\end{figure*}

\subsubsection{Pinpointing default-mode bias with \doesit} 
\label{subsubsec:default_mode_bias} 
We identify cases of over- and under-representation of attributes in images generated with coarse prompts as indicated by high and low, respectively, \doesit scores. % for a given concept. 

We show examples of this analysis in \Cref{fig:default_mode_bias_analysis}. 
For example, we observe that when studying the attribute type \textit{purpose} for images generated with the coarse prompts containing the concept \textit{airplane}, the ``passenger'' attribute has a very high \doesit score across all T2I models, the ``military'' attribute has a very low negative score, and the ``cargo'' and ``private'' attributes have moderate scores. 
Randomly sampled images confirm this observation, showing passenger airplanes by a large margin and few to no military airplanes. 
Similarly, for the concept \textit{table} and attribute type \textit{material}, the ``wooden'' attribute has very high \doesit score while ``glass'' has a very low score. 

Interestingly, we note that the \canit scores of these under-represented attributes are still high: when specifically asked to generate these attributes through dense prompts, the models are able to (last column in \Cref{fig:default_mode_bias_analysis}). 
However, we also find opposite trends.
For example, for the concept \textit{bird} and attribute type \textit{state,} we find that the ``perched'' attribute has high \doesit scores, even though some of the other states, \textit{e.g.} ``flying,'' are also common real-world occurrences. % state with which most birds are associated. 
This result is confirmed when visualizing images in \Cref{fig:default_mode_bias_analysis}.
In ~\Cref{subsec:training_data_investigation} we further explore the relationship between the (explicit or implicit) prevalence of concept-attribute pairs in the training data of the models and their \benchmark scores.

\subsubsection{Finding generalization failures with \canit} 
\label{subsubsec:generalization_failure_modes}
We analyze the \canit to measure generalization capacity. 
As a reminder, a very low negative \canit score for an attribute $a \in A_{c,t}$ of concept $c$ means that even when the model is explicitly prompted to generate the attribute $a$, it generates other attributes of type $A_{c,t}$ more than the specified attribute $a$, indicating a generalization failure. 
We observe these trends in model generalization: 

\textbf{T2I models show strong generalization performance for attribute types like \textit{color}, \textit{material} and \textit{pattern}} across the concepts in \cocobenchmark. \Cref{fig:capacity_analysis} shows an example, where 
\canit scores of all \textit{color} attributes are high, showing that the model can generate colors in images when prompted to do so.
Random examples of images reveal \benchmark appropriately identifies this trend in model capability, shown for the attribute ``blue'' in \Cref{fig:capacity_analysis}.

\textbf{Larger models show strong generalization for some concept-attribute pairs}, with higher positive \canit score while some smaller models have low negative \canit score, such as for ``digital'' \textit{clock}, ``convertible'' \textit{car}, and ``freight'' \textit{train}.
In \cref{fig:capacity_analysis}, we see that the smaller \MetaLDM has a negative \canit for ``convertible'' \textit{car} while the larger models have high, positive scores.
These patterns are reflected in visual inspection, as we see \MetaLDM struggle to generate these attributes in the sampled images when explicitly prompted to do so while \sdthree and \flux show good generalization. However there are a few cases where larger models show weaker generalization.
For example, \flux struggles to show images of a \textit{boat} on ``land,'' while other models are able to do so. 
Examples are shown in the Supplementary Material. 

\textbf{T2I models struggle with negations.} 
For example, \Cref{fig:capacity_analysis} shows that all models have low \canit scores for the concept \textit{bed} when negation attributes ``without blanket'' and ``without pillows,'' generating images of blankets and pillows even when explicitly asked not to generate those. 
This is also observed by previous work~\citep{conwell2024relationsnegationsnumberslooking}, and demonstrates the efficacy of \benchmark in serving as an automatic identifier of established failure modes.

\textbf{Identifying new generalization failure modes}
\benchmark identifies a new failure mode where models have low negative \canit scores but high \doesit scores. 
For example, most models fail to differentiate between ``open'' and ``closed'' attributes when explicitly asked to do so through dense prompts. Sometimes they fail to consistently generate the ``closed'' attribute, while other times they fail to consistently generate ``open'' concepts.
However, models are able to generate both ``open'' and ``closed'' attributes when generating images through coarse prompts. 
Which means that the models fail to generate certain attributes when explicitly asked, but generates them in default-mode. 
Such examples of ``closed-wings'' \textit{birds} generated by \sdthree~\citep{esser2024scalingrectifiedflowtransformers_SD3_5} and ``full'' \textit{refrigerators} generated by \MetaLDM are shown in ~\Cref{fig:surprising_can_it}.
We include more examples in the Supplementary Materials.

\begin{figure}
  \centering
  \includegraphics[width=\linewidth]{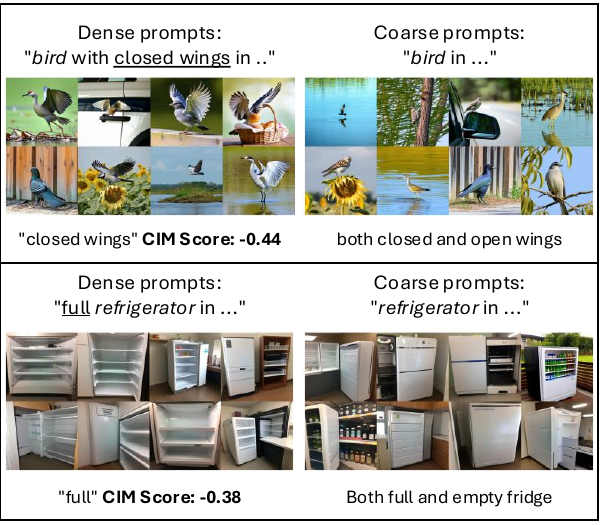}
  \caption{\textbf{\benchmark identifies cases when models struggle to generate an attribute-concept combination when explicitly requested via dense prompts but can with coarse prompts.} 
   For example, \sdthree~\citep{esser2024scalingrectifiedflowtransformers_SD3_5} struggles to generate ``closed-wings'' \textit{birds} when explicitly prompted to, but does with coarse prompts, and \MetaLDM struggles to generate ``full'' \textit{fridges} when prompted to do so, but does with coarse prompts.
  }
  \label{fig:surprising_can_it}
\end{figure}

% -----------------
\subsection{Training data investigation}
\label{subsec:training_data_investigation}

\begin{figure}[h]
    \centering
    \begin{subfigure}[b]{0.49\columnwidth}
        \includegraphics[width=\textwidth]{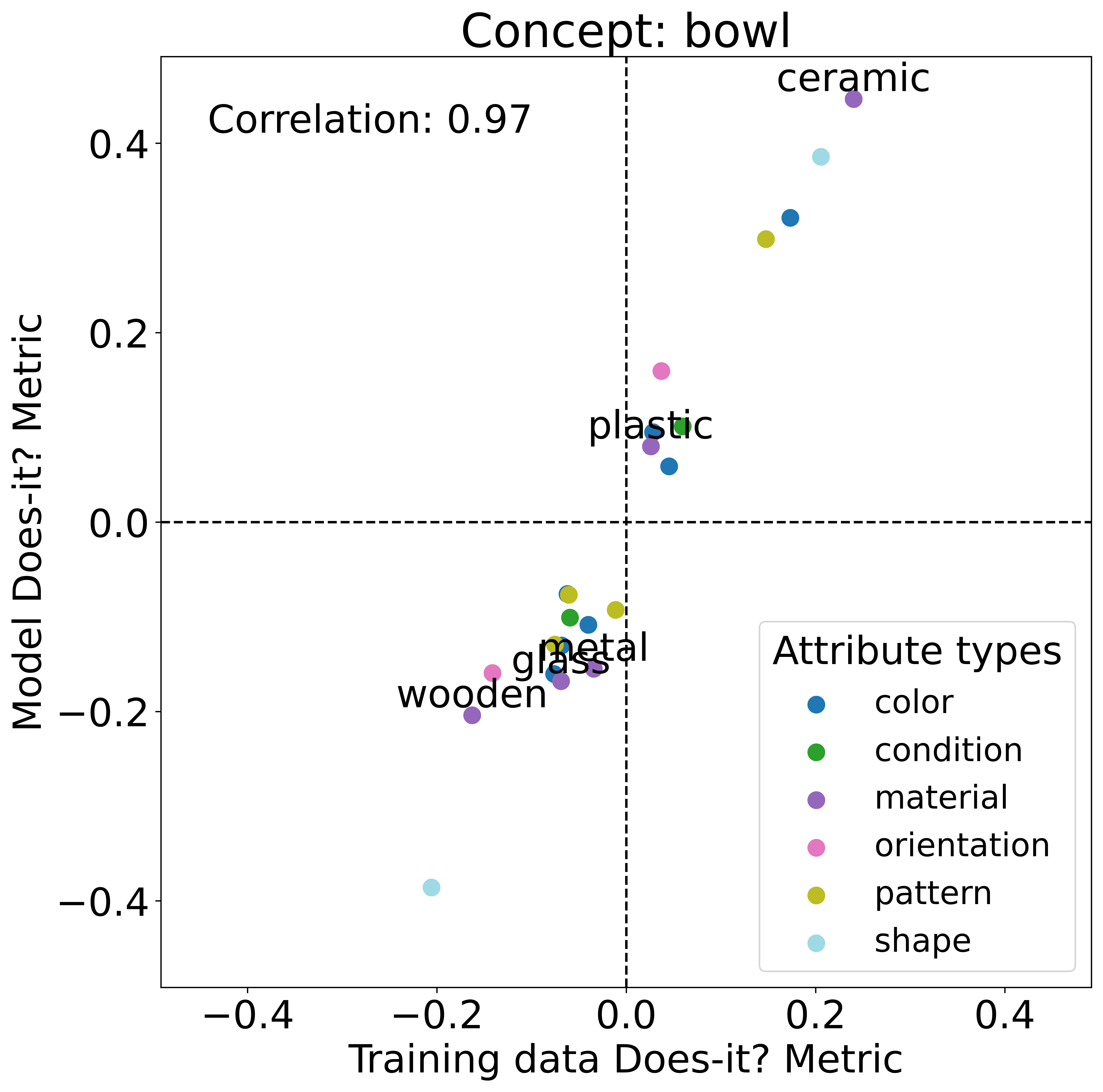}
    \end{subfigure}
    \hfill
    \begin{subfigure}[b]{0.49\columnwidth}
        \includegraphics[width=\textwidth]{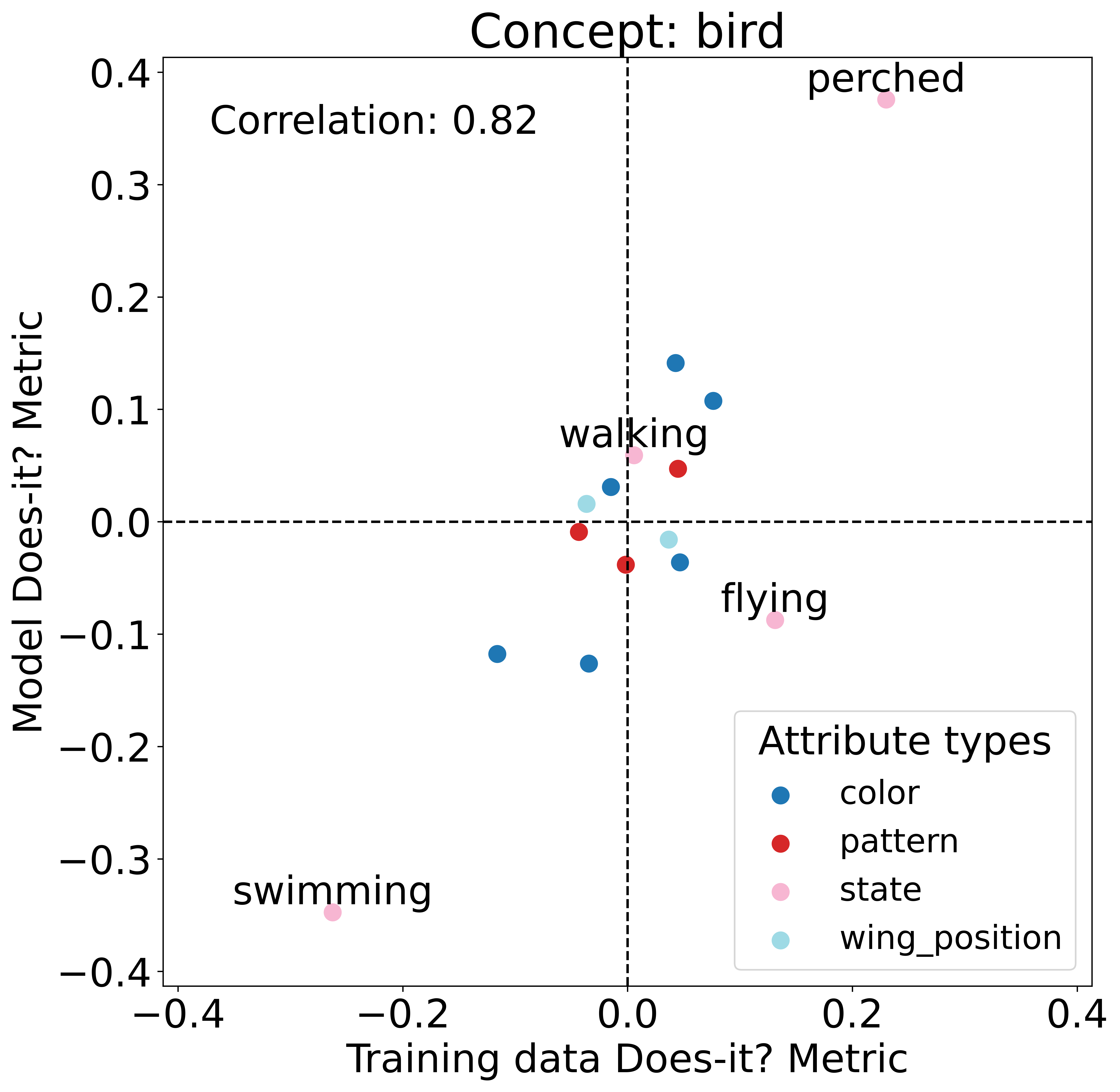}
    \end{subfigure}
    \caption{\textbf{Default-mode model diversity reflects diversity in images used for model training.} We show plots for attributes of concepts \textit{bowl} and \textit{bird} for the \MetaLDM model trained on CC12M~\citep{changpinyo2021cc12m}. The y-axis has  \doesit scores for generated images, and the x-axis has \doesit scores for training images. The scores are highly correlated, although there exist some outliers. More plots can be found in the Supplementary Materials.
    }
    \label{fig:training_data_analysis}
\end{figure}

We investigate the distribution of attributes in training data images to understand possible sources of observed default-mode diversity, as measured by the \doesit scores. 
We use a version of the \MetaLDM model trained on CC12M~\citep{changpinyo2021cc12m} to generate images for coarse prompts in \cocobenchmark and calculate the \doesit for all concept-attribute pairs. 
To find the distribution of concept-attributes in training data images of CC12M~\citep{changpinyo2021cc12m}, we first find images for all concepts in \cocobenchmark by using the concept name as the query text of VQAScore and removing all images that have a score less than 0.8. 
With the remaining images, we calculate the \doesit scores the same way we do for generated images. \looseness=-1

In ~\Cref{fig:training_data_analysis}, we plot the \doesit scores for the generated images and the training data images \doesit scores.
We observe that the two scores are highly correlated. 
For example, ``ceramic'' \textit{bowl} scores high for both the generated and training data images, while the ``wooden'' \textit{bowl} is low. 
The correlation (Pearson correlation coefficient) across  \cocobenchmark for all concept-attribute pairs is 0.85, indicating that the default-mode diversity of attributes in generated images is a close reflection of the skew of those attributes in the training data images.
Furthermore, \benchmark can also be used to identify anomalous cases where attributes have relatively high prevalence in the training data, such as ``flying'' \textit{bird}, but are not generated by the model with coarse prompts (images in \Cref{fig:default_mode_bias_analysis}). 
For these cases, interventions such as increased representation in image training data may not be sufficient to achieve default-mode diversity. 

%% file: sec/6_discussion.tex
\section{Discussion}
\label{sec:discussion}

We introduced a new reference dataset-free benchmarking framework, \benchmark, that is capable of assessing default-mode diversity and generalization capacity of T2I generative models.
Using \benchmark, we identify a trade-off between model generalization and default-mode diversity that becomes evident with increasing model size.
We observe that T2I models easily generate certain uncommon attributes when explicitly prompted but fail to do so in default behavior. 
In terms of sources of default-mode diversity behavior, we utilize the \benchmark to identify correlations between attributes in the training data and attributes generated by using generic attribute-free prompting.
Lastly, we find that \benchmark is capable of finding new model failure cases, such as when particular attributes (\eg, ``open'' or ``closed'') can be generated by attribute-free generic prompts, but not when explicitly requested.

\subsection{Limitations}

\benchmark relies on external models for dataset construction and metric calculation.
While the use of such models in T2I benchmarking has strong precedence~\citep{hessel-etal-2021-clipscore,hu2023tifaaccurateinterpretabletexttoimage_tifa,cho2024davidsonianscenegraphimproving_DSG},
we take steps to address these concerns, using realistic captions to reduce possibility of model hallucinations, filtering attributes to ensure they are mutually exclusive and imageable, and adapting VQAScore to increase reliability.
In addition, because \benchmark uses open source models, it is not subject to uncontrolled variations that would be inherent to a closed-source API. 
Furthermore, the current method does not distinguish between multiple valid definitions of a concept and instead supports a breadth of visual depictions. 
Finally, while \benchmark studies generalization capabilities under different prompts, the model of focus may still be capable of generating concept attributes with methods other than explicit prompting, such as via external guidance ~\citep{hemmat2024improvinggeodiversitygeneratedimages}. \looseness=-1

%% file: sec/7_appendix.tex
% \clearpage
% \onecolumn

% \section{Appendix}
% \label{sec:appendix}

\section{\MetaLDM Training Details}
\label{app:training_details}
 
 The \MetaLDM version evaluated in Section~\ref{sec:experiments_overall} was trained leveraging flow matching~\citep{lipman2023flowmatchinggenerativemodeling_flow_matching}, control conditions~\citep{berrada2025boosting,podell2023sdxlimprovinglatentdiffusion_SDXL} and latent perceptual loss~\citep{berrada2025boosting}.
 The training dataset consisted of image-caption pairs including ImageNet~\citep{5206848}, CC12M~\citep{changpinyo2021cc12m}, YFCC~\citep{DBLP:journals/corr/ThomeeSFENPBL15}, and an internally licensed dataset. 
 The version of \MetaLDM evaluated in Section~\ref{subsec:training_data_investigation} was trained with only CC12M~\citep{changpinyo2021cc12m}.
 
\section{Examples of model generalization capacity analysis through \benchmark}
Here are a few more examples to show how we use \cocobenchmark to find interesting insights about different models' generalization capacity.

\begin{figure}[h]
  \centering
  \includegraphics[width=0.9\textwidth]{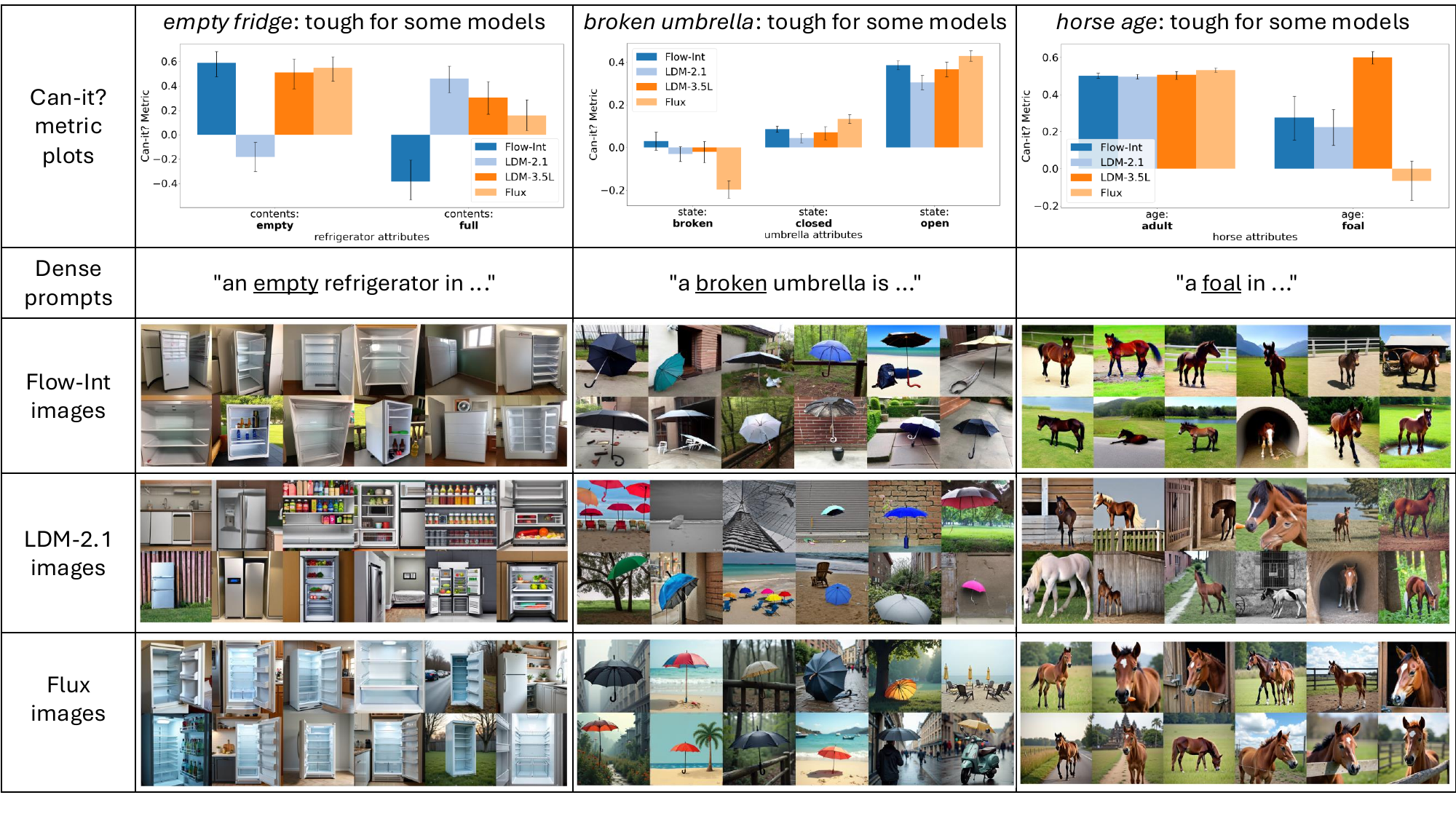}
  \caption{
  \textbf{Interesting generalization failure modes found through \cocobenchmark:}
  \textbf{[Left]} \sdtwo~\cite{Rombach_2022_CVPR_SD2_1} struggles to generate empty refrigerators even when prompted to do so. 
  \textbf{[Middle]} Most models struggle to generate  ``broken'' \textit{umbrellas}, but \flux~\cite{flux2024} is especially poor at it. \flux has a very low negative \canit for ``broken'' and most of its generated images have completely unbroken umbrellas as seen above. \textbf{[Right]} Smaller models (\MetaLDM and \sdtwo) are better at generating young \textit{horses}, \textit{e.g.} ``foals'', than \flux (as observed through the \canit and sampled images). 
  }
  \label{fig:app_interesting_failure_modes_2}
\end{figure}

\begin{figure}[h]
  \centering
  \includegraphics[width=0.9\textwidth]{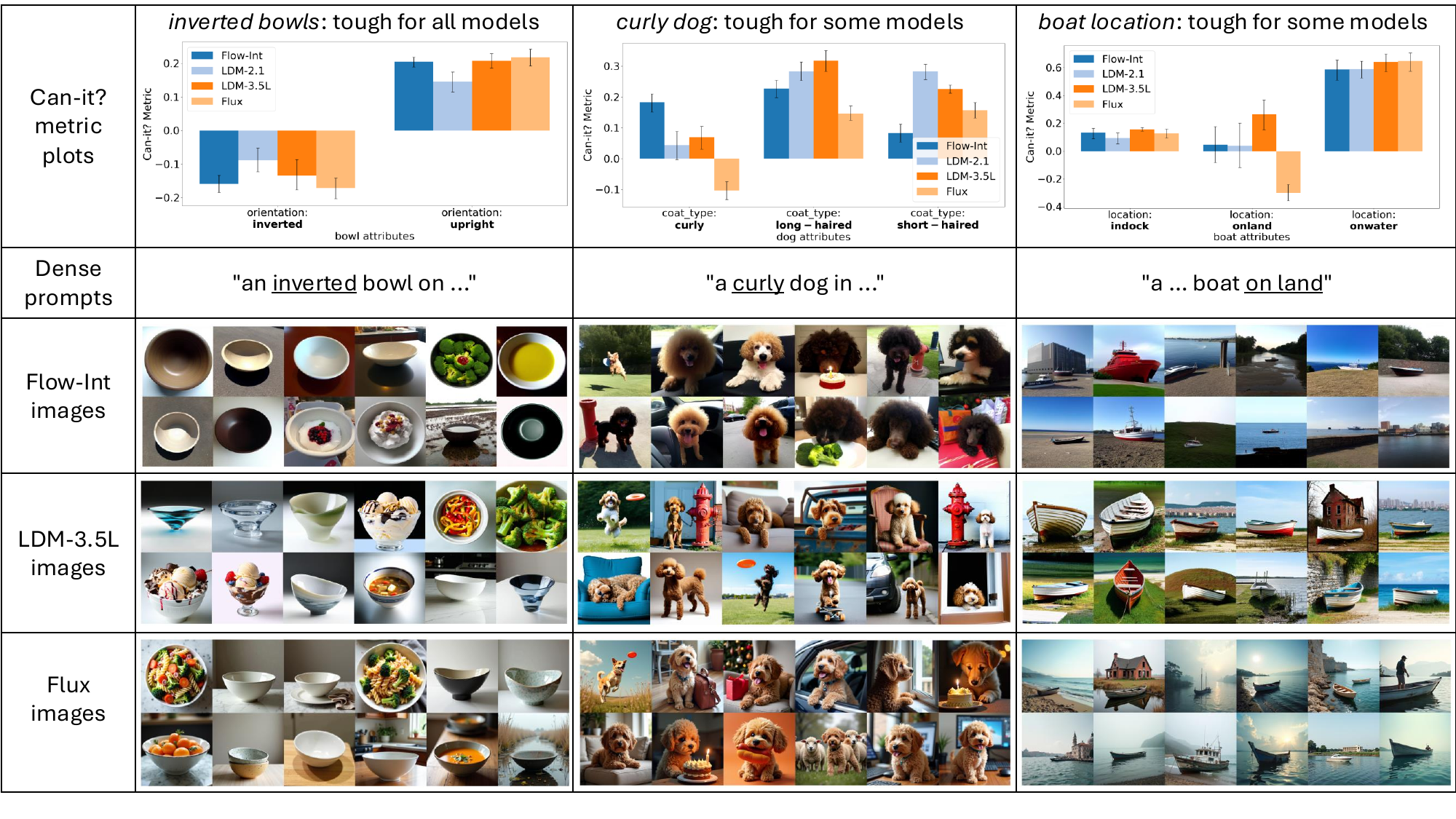}
  \caption{
  \textbf{Interesting generalization failure modes found through \cocobenchmark:} 
  \textbf{[Left]} We find that none of the models are able to generate ``inverted'' \textit{bowls}. \textbf{[Middle]} \flux~\cite{flux2024} is not good at generating \textit{dogs} with ``curly'' \textit{hair}, even though other models which are much smaller than \flux are able to. 
  \textbf{[Right]} \flux is also bad at generating \textit{boats} on ``land''. Other smaller models are much better at generating such cases.
  }
  \label{fig:app_interesting_failure_modes_1}
\end{figure}

\begin{figure}[h]
  \centering
  \includegraphics[width=0.9\textwidth]{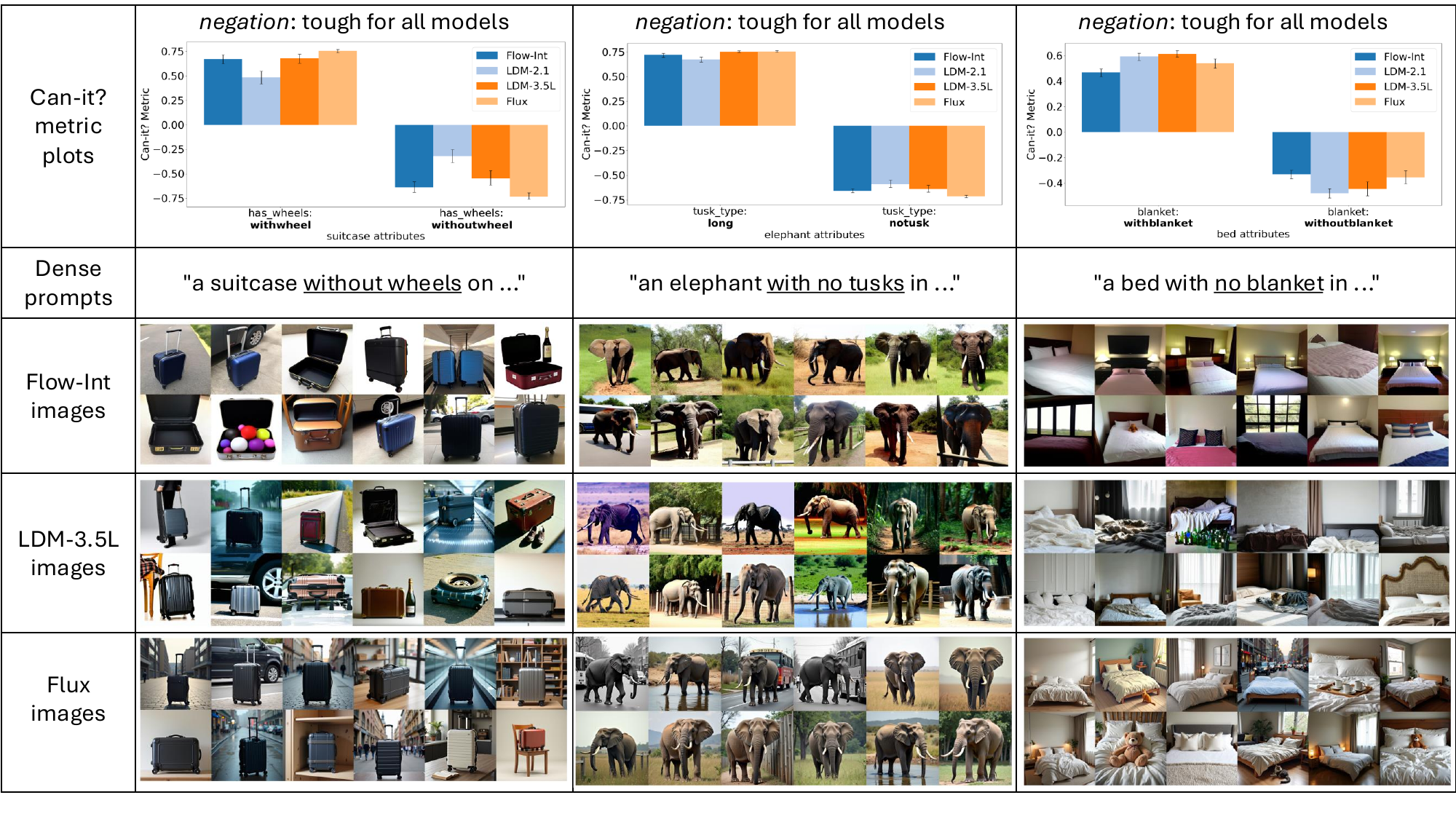}
  \caption{
  \textbf{Most T2I models are not able to handle negation in prompts.} Models generate the negated attributes in the prompts. The dense prompts in the first column say ``without wheels" but all the models generate suitcase wheels in most of the generated images. Similarly, they generate elephant tusks and blankets on beds in most of the generated images, even though the prompts specifically asks them not to. 
  }
  \label{fig:app_negation_tough}
\end{figure}

\begin{figure}[h]
  \centering
  \includegraphics[width=0.9\textwidth]{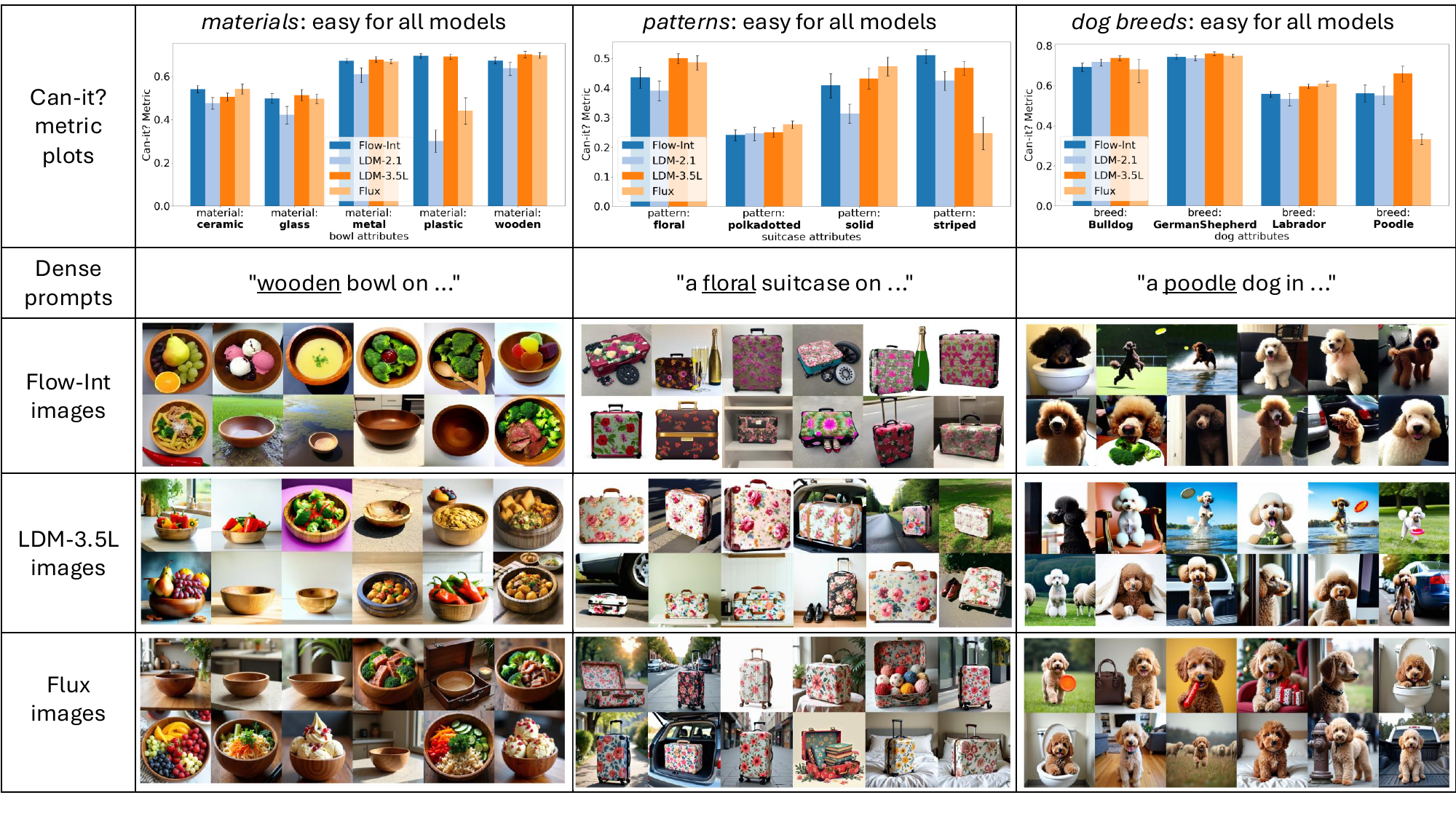}
  \caption{\textbf{Most models are good with attributes like color, pattern, material and breeds:} Examples showing that \canit for these concept-attributes are high and positive for most models. Generated images also show that the models can easily generalize to these attributes.}  
  \label{fig:app_easy_attributes}
\end{figure}

\begin{itemize}
    \item Figure~\ref{fig:app_interesting_failure_modes_1} and Figure~\ref{fig:app_interesting_failure_modes_2} has examples that show some unique failure modes found through \cocobenchmark. In some cases larger size models struggle to generate an attribute that is easily generated by smaller size models (like ``broken'' \textit{umbrella} and ``foal'' in Figure~\ref{fig:app_interesting_failure_modes_2} and  ``on land'' \textit{boat} location and ``curly'' \textit{dog} in Figure~\ref{fig:app_interesting_failure_modes_1}). While in some cases, all of the models struggle (``inverted'' \textit{bowls} in Figure~\ref{fig:app_interesting_failure_modes_1}), in a peculiar case, \sdtwo~\cite{Rombach_2022_CVPR_SD2_1} struggles to generate ``empty'' \textit{refrigerators} in Figure~\ref{fig:app_interesting_failure_modes_2}.

    \item Figure~\ref{fig:app_negation_tough} has examples that show that negations are difficult for all models. When these models are prompted to not to generate an attribute, more often than not they generate those attributes in the images.

    \item  Figure~\ref{fig:app_easy_attributes} has examples that show that some attribute types like \textit{material}, \textit{patterns} and \textit{dog breeds} are easy for all models. All the models have high \canit for attributes of these types and can generate these attributes in images when prompted. 
\end{itemize}

\clearpage
\section{Example plots to show that model default-mode diversity reflects training data diversity}
\begin{figure}[h]
    \centering
    \begin{subfigure}[b]{0.32\columnwidth}
        \includegraphics[width=\textwidth]{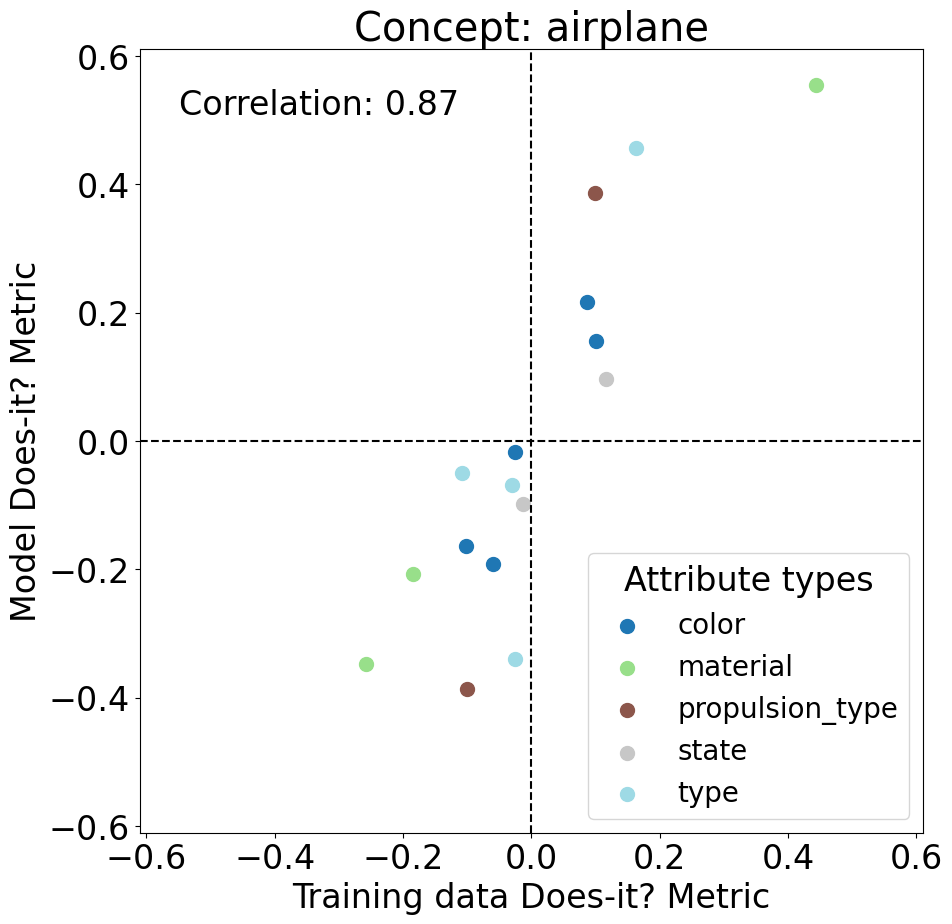}
    \end{subfigure}
    \hfill
    \begin{subfigure}[b]{0.32\columnwidth}
        \includegraphics[width=\textwidth]{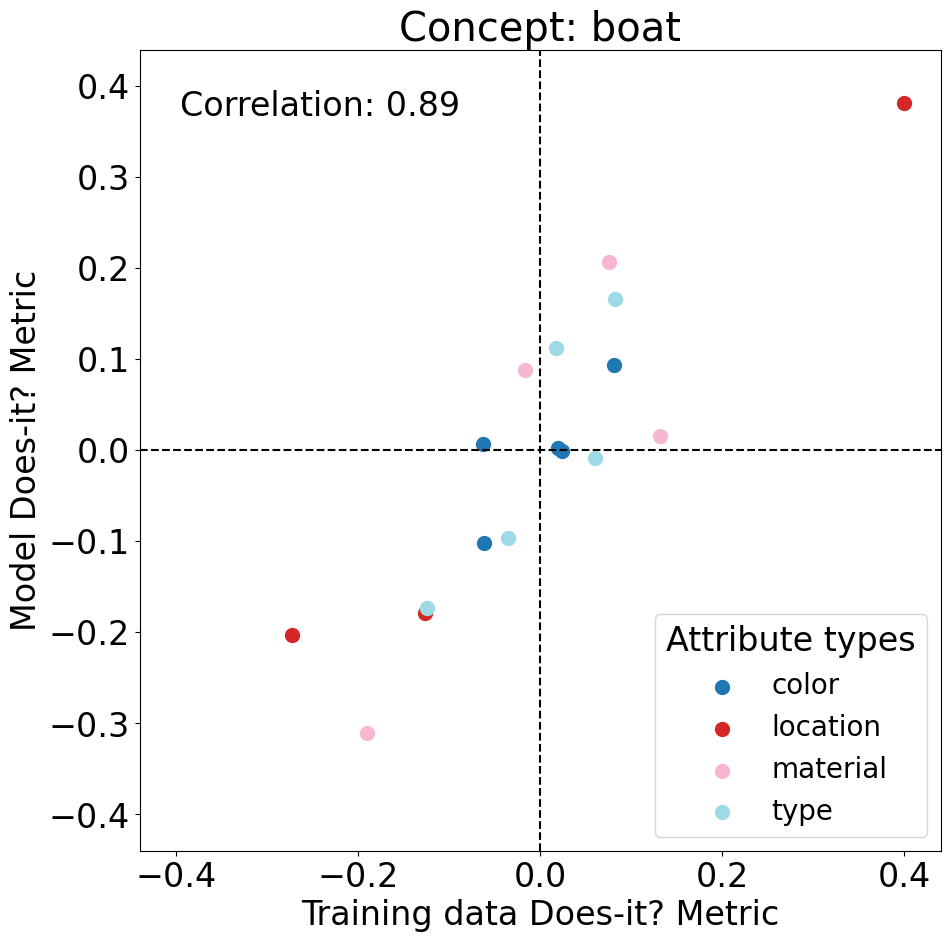}
    \end{subfigure}
    \hfill
    \begin{subfigure}[b]{0.32\columnwidth}
        \includegraphics[width=\textwidth]{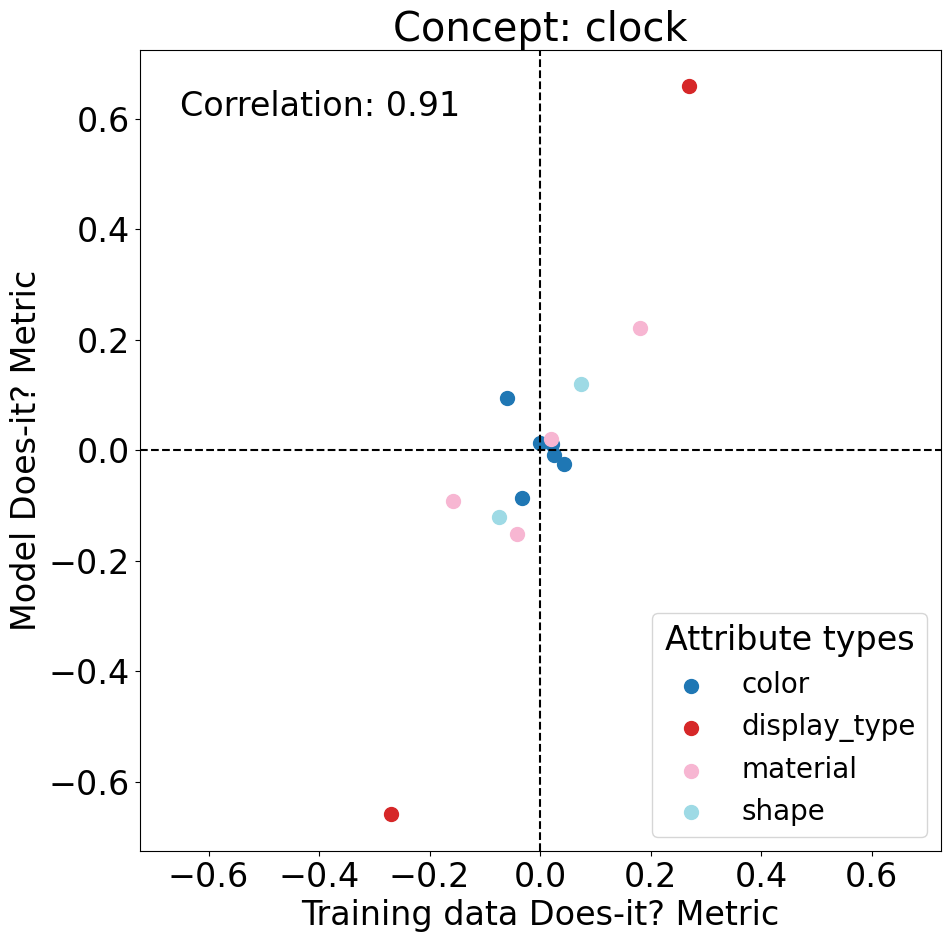}
    \end{subfigure}
    \vspace{0.1cm} % add some vertical space
    \begin{subfigure}[b]{0.32\columnwidth}
        \includegraphics[width=\textwidth]{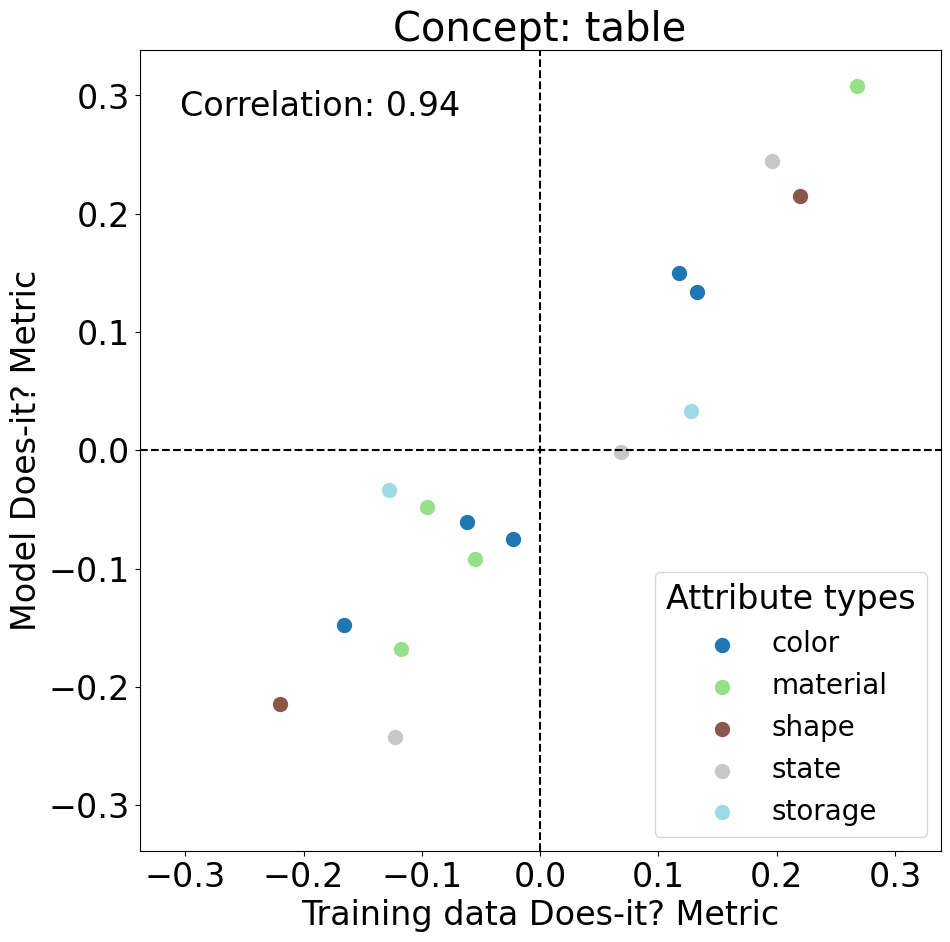}
    \end{subfigure}
    \hfill
    \begin{subfigure}[b]{0.32\columnwidth}
        \includegraphics[width=\textwidth]{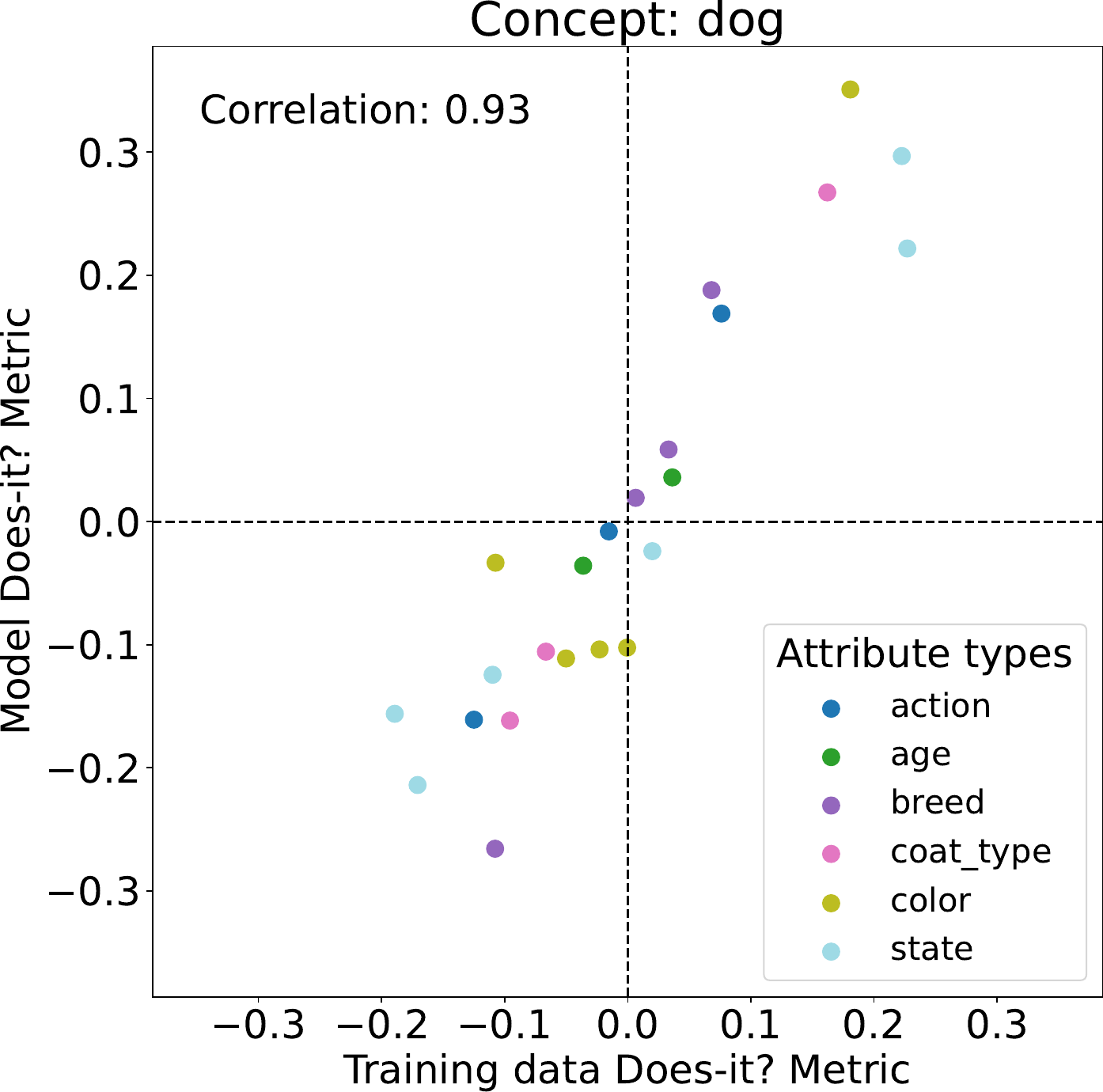}
    \end{subfigure}
    \hfill
    \begin{subfigure}[b]{0.32\columnwidth}
        \includegraphics[width=\textwidth]{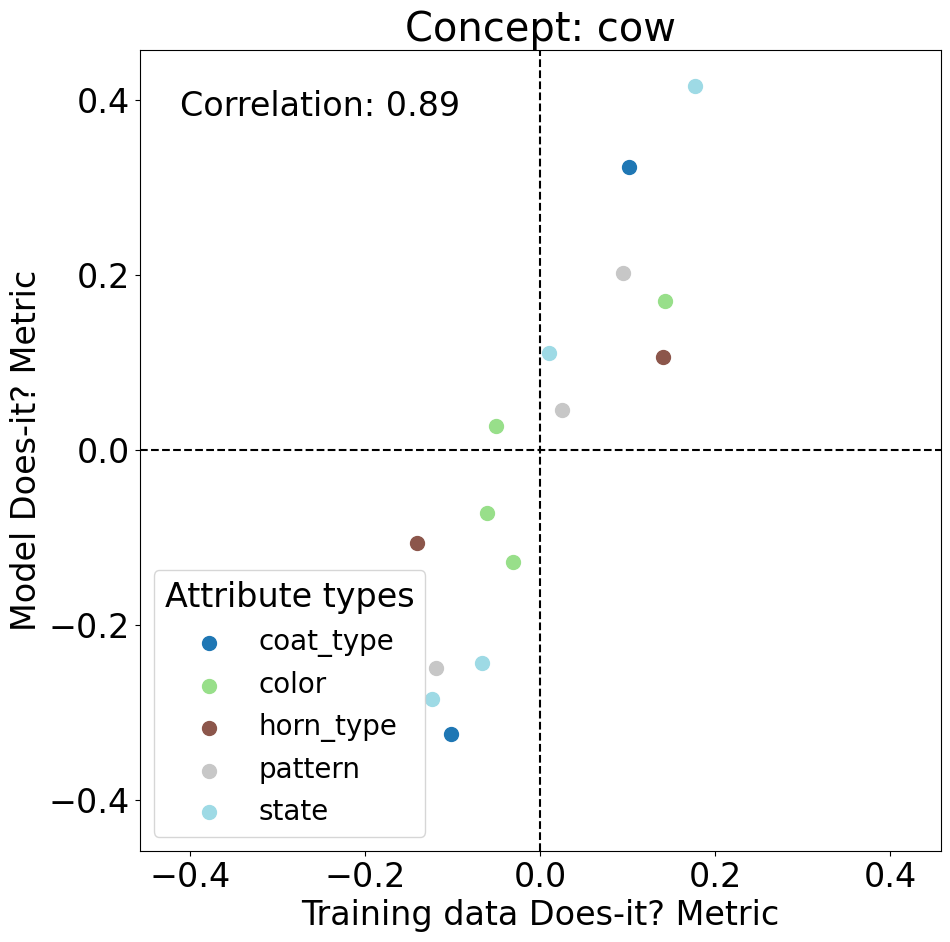}
    \end{subfigure}
    \caption{\textbf{Default-mode model diversity reflects diversity in images used for model training}: We show plots for attributes of 6 concepts (airplane, boar, clock, cow, dog, and table). On the y-axis we have \doesit scores calculated from generated images of the concepts and on the x-axis we have \doesit scores calculated from the training data images of those concepts. We see that the scores are highly correlated.
    }
    \label{fig:app_training_data_analysis_scatter_plots}
\end{figure}

In Figure~\ref{fig:app_training_data_analysis_scatter_plots}, we plot the \doesit scores for the generated images for a few more concepts and the \doesit scores of their corresponding training data images. The \doesit scores are highly correlated and show that model default-mode diversity reflects training data diversity.

\clearpage
\section{\cocobenchmark dataset creation figure}
\begin{figure}[h!]
  \centering
  \includegraphics[width=\textwidth]{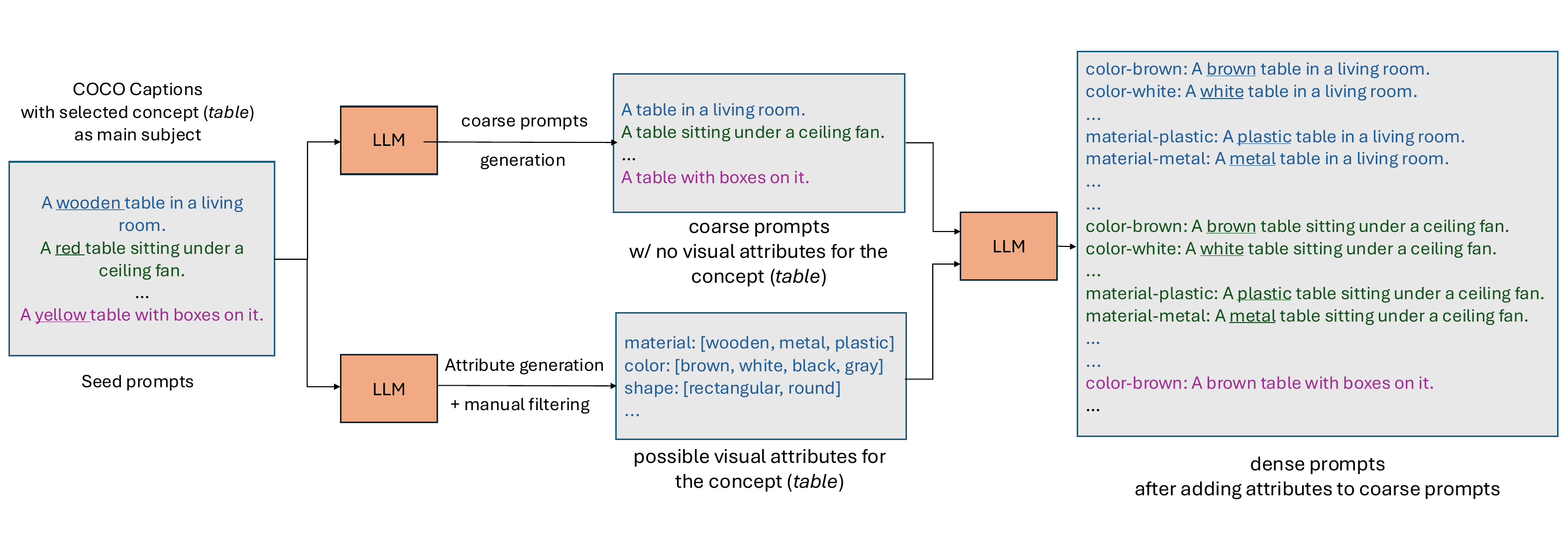}
  \caption{\textbf{\cocobenchmark dataset creation:} For a concept (\textit{table} in this example), we start by randomly selecting captions from COCO~\cite{lin2015microsoftcococommonobjects_coco} as seed prompts. We then use an LLM (Llama3.1~\cite{grattafiori2024llama3herdmodels}) to generate coarse prompts from seed prompts and find possible concept attributes. To generate dense prompts from a coarse prompt, we use the LLM (Llama3.1~\cite{grattafiori2024llama3herdmodels}) to inject attributes from the possible visual attributes to the coarse prompt, thus creating multiple dense prompts with different visual attributes.
  }
  \label{fig:app_dataset_creation}
\end{figure}

Figure~\ref{fig:app_dataset_creation} shows a flow chart of \cocobenchmark dataset creation process. It shows how we use LLMs to collect attributes, coarse prompts and dense prompts from seed prompts - as explained in Section~\ref{sec:benchmark_dataset}.

\clearpage
\section{Meta-prompts for instructing the LLM}
\label{app:llama_meta_prompts}
We instruct LLMs (Llama3.1~\citep{grattafiori2024llama3herdmodels}) to generate \cocobenchmark. More specifically, we instruct Llama3.1 to (1) create a list of potential attributes of a concept from a given seed prompt and (2) generate coarse prompts and dense prompts from seed prompts and list of attributes. The Python functions to generate LLM instruction prompts are given in Table~\ref{fig:llama_prompt_to_generate_attributes} and Table~\ref{fig:llama_prompt_to_generate_coarse_dense_prompts}.

\lstdefinestyle{mystyle}{
    numbers=none,
    breaklines=true,
    breakindent=0pt,
    breakautoindent=false,
    showstringspaces=false,
    basicstyle=\fontsize{7}{7}\ttfamily,
    rulecolor=\color{black},
    moredelim=**[is][\color{teal}]{<SYS>}{</SYS>},
    moredelim=**[is][\color{purple}]{<FILL>}{</FILL>},
    moredelim=[is][\color{blue}]{**}{**},
}

\begin{table*}[ht]
\centering
\caption{Meta-prompt for Llama to generate a list of potential attributes of a concept from a given \texttt{seed\_prompt} used for image generation.}
\label{fig:llama_prompt_to_generate_attributes}

\begin{tabular*}{\linewidth}{@{\extracolsep{\fill}} l }
\begin{lstlisting}[frame=single, style=mystyle]
Given an image caption, find the main subject of the caption.
Once you find the main subject of the caption, 
find the visual modifiers or visual attributes of the main subject in the caption. 
Give a list of visual modifier/attribute types. If there are no modifiers/attributes, give an empty list
For example, in this example caption
c: "a black dog running on the beach"
the main subject is 'dog'
visual modifiers types and values are:
color: black
state: running

Now can you find main subject and visual modifier/attribute types and values for this caption? 
Please ignore the context and scene related attributes/modifiers.
c = <FILL>{seed_prompt}</FILL>

what are some of the other possible attribute/modifier types for the above main subject?
Also, what are some of the possible values for those attributes/modifiers? 
Please find attributes relevant to the main subject and the caption.
can you answer in a nice json format? 
Put all the existing visual attribute types and other possible attribute types in the json with their possible values.

For example, the output json for the dog example above is:

{
    "caption": "a black dog running on the beach",
    "main_subject": "dog",
    "visual_modifiers": {
        "existing": {
        "color": "black",
        "state": "running"
        },
        "possible_attributes": {
        "color": ["black", "white", "brown", "gray", "golden"],
        "breed": ["Labrador", "German Shepherd", "Poodle", "Bulldog"],
        "size": ["small", "medium", "large"],
        "age": ["puppy", "adult", "senior"],
        "coat_type": ["short-haired", "long-haired", "curly"],
        "body_type": ["muscular", "slim", "stocky"],
        "state": ["running", "sitting", "standing", "lying down", "jumping"]
        }
    }
}

Please output similar json for c = <FILL>{seed_prompt}</FILL> with possible attribute/modifier types and their possible values. 
Output only the json and nothing else.
Output: 
\end{lstlisting} \\
\end{tabular*}
\end{table*}

\begin{table*}[ht]
\centering
\caption{Meta-prompt for LLama used to generate coarse prompts and dense prompts from an input file \texttt{attributes\_json} that contains seed (image generation) prompt and potential visual attributes.}
\label{fig:llama_prompt_to_generate_coarse_dense_prompts}

\begin{tabular*}{\linewidth}{@{\extracolsep{\fill}} l }
\begin{lstlisting}[frame=single, style=mystyle]
<FILL>{attributes_json}</FILL>

given the above json as input_json

first create a seed prompt by removing all the visual modifiers from the caption. 
Keep the main subject and contextual/environment related attributes as the original prompt.

Once you get the seed prompt, select an attribute type and a value from the json and modify the seed prompt to add those - call it a dense prompt
Do these for all the attribute types and their values to create dense prompts

Make sure that the dense prompts are plausible captions of naturally occuring images. 
If it does not seem naturally plausible, skip that attribute value to create dense prompt

Give output in a nice json format indicating the original caption, seed prompt, selected attribute type, 
selected attribute value and the generated dense prompt after adding the selected attribute

The structure of json should like this example:
{   "original_caption": "a black dog running on the beach",
    "seed_prompt": "a dog on the beach",
    "main_subject": "dog",
    "modified_prompts": [
      {
        "attribute_type": "color",
        "attribute_value": "white",
        "generated_prompt": "a white dog on the beach"
      },
      {
        "attribute_type": "color",
        "attribute_value": "brown",
        "generated_prompt": "a brown dog on the beach"
      },
      .
      .
      .
    ]
}

Output only the json with the above example fields and nothing else. 
Make sure you include all the attribute types and their values from the input_json to create dense prompts

Output: 
\end{lstlisting} \\
\end{tabular*}
\end{table*}